\documentclass{egpubl}
\usepackage{egsgp2025}
\SpecialIssuePaper

\CGFccby

\usepackage[T1]{fontenc}
\usepackage{dfadobe}  
\usepackage{booktabs}

\usepackage{multirow}
\usepackage{epsfig}
\usepackage{graphicx}
\usepackage{amsmath}
\usepackage{amssymb}
\usepackage{overpic}
\usepackage{enumitem}
\usepackage{layouts}
\usepackage{hyphenat,balance,microtype}
\usepackage[fleqn,tbtags]{mathtools}
\usepackage[detect-weight]{siunitx}
\usepackage{caption, subcaption, textpos}
\usepackage{colortbl}
\usepackage{soul}
\usepackage{color, xcolor}

\usepackage{algorithm}
\usepackage{algorithmic}
\usepackage{verbatim}
\usepackage{listings}
\usepackage{wrapfig}
\biberVersion
\BibtexOrBiblatex
\usepackage[backend=biber,bibstyle=EG,citestyle=alphabetic,backref=true]{biblatex} 
\electronicVersion
\PrintedOrElectronic

\usepackage{egweblnk}  

\usepackage{booktabs}
\newcommand{\tablestyle}[2]{\setlength{\tabcolsep}{#1}
                            \renewcommand{\arraystretch}{#2}
                            \centering
                            \footnotesize}

\title[OctFusion]{OctFusion: Octree-based Diffusion Models ~\\ for 3D Shape Generation}

\author[Bojun Xiong \& Si-Tong Wei \& Xin-Yang Zheng \& Yan-Pei Cao \& Zhouhui Lian \& Peng-Shuai Wang]
{
\parbox{\textwidth}{\centering Bojun Xiong\thanks{Denotes equal contributions.}$^{1}$\orcid{0009-0009-6460-8393}, 
Si-Tong Wei\footnotemark[1]$^1$\orcid{0000-0002-8215-6142}, 
Xin-Yang Zheng$^2$\orcid{0000-0003-2318-1863},
Yan-Pei Cao$^3$\orcid{0000-0002-0416-4374},
Zhouhui Lian$^1$\orcid{0000-0002-2683-7170}
and Peng-Shuai Wang\thanks{Corresponding author. E-mail: wangps@hotmail.com}$^1$ \orcid{0000-0001-9700-8188}
}
\\
{\parbox{\textwidth}{\centering $^1$ Peking University, P.~R~China, \qquad $^2$ Tsinghua University, P.~R~China, \qquad $^3$ VAST}}
}

\begin{document}

\teaser{
 \includegraphics[width=0.9\linewidth]{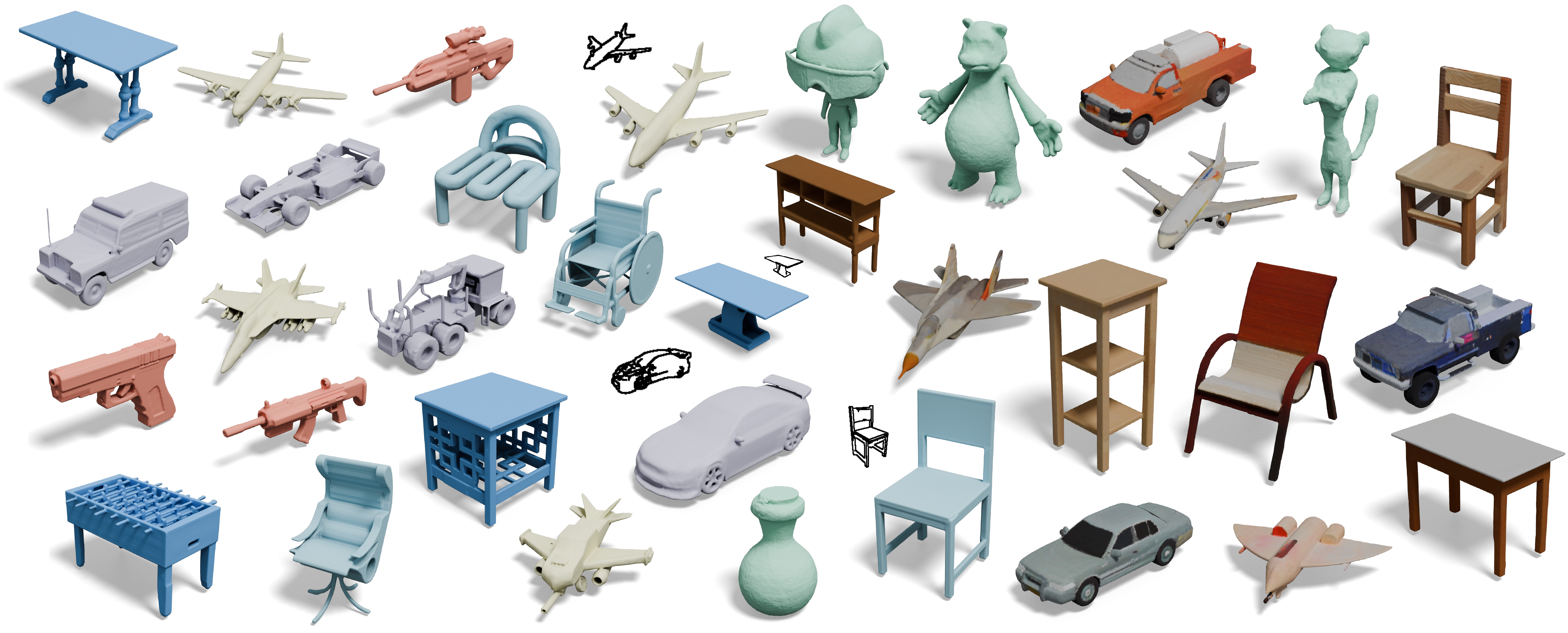}
 \centering
  \caption{OctFusion is capable of generating high-quality and high-resolution 3D shapes in various scenarios, such as unconditional/label-conditional generation, text/sketch-guided generation, and textured mesh generation.}
\label{fig:teaser}
}

\maketitle

\begin{abstract}
Diffusion models have emerged as a popular method for 3D generation.
However, it is still challenging for diffusion models to efficiently generate diverse and high-quality 3D shapes.
In this paper, we introduce OctFusion, which can generate 3D shapes with arbitrary resolutions in 2.5 seconds on a single Nvidia 4090 GPU, and the extracted meshes are guaranteed to be continuous and manifold.
The key components of OctFusion are the octree-based latent representation and the accompanying diffusion models.
The representation combines the benefits of both implicit neural representations and explicit spatial octrees and is learned with an octree-based variational autoencoder.
The proposed diffusion model is a unified multi-scale U-Net that enables weights and computation sharing across different octree levels and avoids the complexity of widely used cascaded diffusion schemes.
We verify the effectiveness of OctFusion on the ShapeNet and Objaverse datasets and achieve state-of-the-art performances on shape generation tasks.
We demonstrate that OctFusion is extendable and flexible by generating high-quality color fields for textured mesh generation and high-quality 3D shapes conditioned on text prompts, sketches, or category labels.
Our code and pre-trained models are available at \url{https://github.com/octree-nn/octfusion}.


\end{abstract}  

\section{Introduction} \label{sec:intro}
3D content creation is a fundamental task in computer
graphics and has a broad range of applications, such as virtual reality, augmented reality, 3D games, and movies.
Recently, generative neural networks, especially diffusion models~\cite{Dickstein2015,Ho2020}, have achieved remarkable progress in 3D generation and have attracted much attention in academia and industry.

However, it is still challenging for diffusion models to efficiently generate highly detailed 3D shapes.
Several works seek to generate 3D shapes by distilling multiview information contained in well-trained 2D diffusion models~\cite{Poole2022,Lin2023}, which involves a \emph{costly} per-shape optimization process and requires minutes, even hours, to generate one single 3D output.
Moreover, the generated results often suffer from the multi-face Janus problem and may contain oversaturated colors.
On the other hand, many works~\cite{Zheng2023,Cheng2023,Hui2022,Li2023,Chou2023,Zhang2023a,Gupta2023} directly train diffusion models on 3D shape datasets.
Although these methods can generate 3D shapes in several seconds by directly forwarding the trained 3D models, the results are often of low resolution due to the limited expressiveness of shape representations like Triplanes~\cite{Gupta2023,Shue2023} or high computational and memory cost of 3D neural networks.
To generate fine-grained geometric details, multi-stage latent diffusion models~\cite{Rombach2022,Cheng2023,Zhang2023a} or cascaded training schemes~\cite{Zheng2023,Ren2024} have been introduced, which further increases the complexity of the training process.

The key challenges for an efficient 3D diffusion model include how to efficiently represent 3D shapes and how to train the associated diffusion models.
In this paper, we propose octree-based latent representations and a unified multiscale diffusion model, named OctFusion, to address these challenges, contributing an efficient 3D diffusion scheme that can generate 3D shapes with effective resolution of up to $1024^3$ in a feed-forward manner.

For the shape representation, we represent a 3D shape as a volumetric octree and append a latent feature on each leaf node.
The latent features are decoded into local signed distance fields (SDFs)  with a shared MLP, which are then fused into a global SDF by multi-level partition-of-unity (MPU) modules~\cite{Wang2022,Ohtake2003}.
This representation combines the benefits of both implicit representations~\cite{Chen2019,Park2019,Mescheder2019} for representing continuous fields and explicit spatial octree structures~\cite{Wang2022} for expressing complex geometric and texture details.
The octree can be constructed from a point cloud or a mesh by recursive subdivision;
the latent features are extracted with a variational autoencoder (VAE)~\cite{Kingma2013} built upon dual octree graph networks~\cite{Wang2022}.
The representation can also be extended to support color fields by additional latent features.
The SDF and color fields can be converted to triangle meshes paired with an RGB color on each vertex as output with the Marching Cubes algorithm~\cite{Lorensen1987}. Although some previous methods also utilized octree~\cite{Zheng2023} or sparse voxel~\cite{Ren2024, xiang2024structured} to achieve high-resolution 3D shape generation, their representation can not guarantee the completeness and continuity of the generated implicit field. On the contrary, our dual octree graph representation forms a complete coverage of the 3D bounding volume, which is further discussed in Section~\ref{sec:ablation}.

\looseness=-1

To train diffusion models on the octree-based representation, our key insight is to regard the splitting status of octree nodes as 0/1 signals;
then, we add noise to both the splitting signal and the latent feature defined on each octree node to get a noised octree.
The diffusion model essentially trains a U-Net~\cite{Ronneberger2015} to revert the noising process to predict clean octrees from noised octrees for generation.
Since the noise is added to all octree levels, a natural idea is to adopt cascaded training schemes~\cite{Zheng2023,Ren2024} to train a \emph{separate} diffusion model on each octree level to predict the splitting signals and the final latent features in a coarse-to-fine manner.
However, training multiple diffusion models is complex and inefficient, especially for deep octrees.
Our key observation is that the octree itself is hierarchical;
when generating the deep octree nodes, the shallow nodes have already been generated, resulting in nested U-Net structures for different octree levels.
Based on this observation, we propose to train a \emph{unified} diffusion model for different octree levels, which reuses the trained weights for shallow octree levels nodes when denoising deep octree nodes.
Our diffusion model enables weights and computation sharing across different octree levels, thus significantly reducing the parameter number and the training complexity and making our model capable of generating detailed shapes efficiently.

We verify the efficiency, effectiveness, and generalization ability of OctFusion on the widely-used ShapeNet dataset~\cite{Chang2015}.
OctFusion achieves state-of-the-art performances with only 33M trainable parameters.
The generated implicit fields are guaranteed to be continuous and can be converted to meshes with arbitrary resolutions.
OctFusion can predict a mesh in less than 2.5 seconds on a single Nvidia 4090 GPU under the setting of 50 diffusion sampling steps.
We also train OctFusion on a subset of Objverse~\cite{Deitke2023} and verify it has a strong  ability to generate shapes from a complex distribution.
We further extend OctFusion to support conditioned generation from sketch images, text prompts, and category labels, where OctFusion also achieves superior performances compared with previous methods~\cite{Zheng2023,Cheng2023}.
In summary, our main contributions are as follows:
\begin{itemize}
\item[-] We present an octree-based latent representation for high resolution 3D shapes modeling, which combines the benefits of both implicit representations and explicit spatial octree structures.
\item[-] We designed a unified multi-scale 3D diffusion model that can efficiently synthesize high-quality 3D shapes in a feed-forward manner within 2.5 seconds.
\item[-] Our proposed OctFusion demonstrates the state-of-the art performances on unconditional generation and conditoned on text prompts, sketch images or category labels. Extensive experiments have been conducted on these tasks to verify the superiority of our method over other existing approaches, indicating its effectiveness and broad applications.

\end{itemize}

\section{Related Work} \label{sec:related}

\subsection{3D Shape Representations}
Different from images that are often defined on regular grids, 3D shapes have different representations.
Early works~\cite{Wu2015,Wu2016,Zheng2022} represent 3D shapes as uniformly sampled voxel grids, with which image generative models can be directly extended to the 3D domain.
However, voxel grids incur huge computational and memory costs; thus, these methods can only generate 3D shapes with low resolutions.
To improve the efficiency, sparse-voxel-based representations, like octrees~\cite{Wang2018a,Wang2022} or Hash tables~\cite{Choy2019,Muller2022} are proposed to represent 3D shapes with only non-empty voxels, which enables the generation of high-resolution 3D shapes.
Another type of 3D representation is point clouds.
Due to the flexibility and efficiency of point clouds, they are widely used for 3D generation~\cite{Fan2017,Lou2021,Liu2019a,Nichol2022,Zeng2022}.
However, point clouds are discrete and unorganized; thus, additional efforts are needed to convert point clouds to continuous surfaces, which are more desirable for many graphics applications.
To model continuous surfaces of 3D shapes, MLP-based distance or occupancy fields are proposed as implicit representations of 3D shapes~\cite{Park2019,Chen2019,Mescheder2019}.
Although these methods can represent 3D shapes with infinite resolutions, they are computationally expensive since each query of the fields requires a forward pass of the MLP.
Recently, triplanes~\cite{Peng2020,Shue2023} are combined with MLPs to further increase the expressiveness and efficiency, whereas it is observed that triplanes are still hard to model complex geometric details~\cite{Zhang2023a,Wang2022,Zhang2022a}.
Our shape representation extends the neural MPU in~\cite{Wang2022} and combines the benefits of both implicit representations and octrees, which can represent continuous fields and model complex geometric and texture details efficiently.

\subsection{3D Diffusion Models}
Recently, diffusion models~\cite{Dickstein2015,Ho2020} have demonstrated great potential for generating diverse and high-quality samples in the image domain \cite{Ramesh2022,Saharia2022,Rombach2022}, surpassing the performance of GANs~\cite{Goodfellow2016a,Dhariwal2021} and VAEs~\cite{Kingma2013}.
Following this progress, a natural idea is to extend them to the 3D domain.
3D diffusion models with point clouds~\cite{Nichol2022,Lou2021,Zhou2021a} or voxel grids~\cite{Li2023,Hui2022,Chou2023,Shim2023} are first proposed and have achieved promising shape generation results.
However, the efficiency and quality of generated shapes are relatively low.
Inspired by latent diffusion models~\cite{Rombach2022}, many follow-up works also train diffusion models on the latent space of 3D shapes, and the latent space is often obtained with a VAE trained on voxels~\cite{Cheng2023}, point clouds~\cite{Zeng2022}, triplanes~\cite{Gupta2023,Shue2023}, or implicit shape representations~\cite{Zhang2023a,Jun2023,Erkoc2023}.
Another strategy for efficiency and quality improvement is to leverage sparse-voxel-based representations, like octree~\cite{Zheng2023}, and adopt cascaded training schemes to generate sparse voxels at different resolutions~\cite{Zheng2023}.
Subsequent works continue to train cascaded models on the larger dataset~\cite{Liu2023, xiang2024structured} or increase the number of training stages for higher resolutions~\cite{Ren2024}.
Different from these methods, our OctFusion is trained with an unified U-net and can generate 3D shapes with $1024^3$ resolutions in a single network, which significantly reduces the complexity of the training procedure.
\looseness=-1

\subsection{3D Generation with 2D Diffusion Priors}
Apart from training 3D diffusion models, another line of research attempts to distill multi-view image priors from 2D diffusion models to generate 3D shapes, popularized by DreamFusion~\cite{Poole2022} and Magic3D~\cite{Lin2023}.
The key idea is to optimize 3D representations such as NeRF~\cite{Mildenhall2020} or InstantNGP~\cite{Muller2022} with the gradient guidance from 2D diffusion priors~\cite{Poole2022,Wang2023a}.
Although these methods can generate diverse and high-quality 3D shapes without access to 3D training data, they are computationally expensive since the optimization of 3D representations often takes minutes or even hours;
and the resulting textures are often oversaturated, and the generated shapes contain artifacts due to the inconsistency of multiview image priors provided by 2D diffusion models.
Many follow-up works~\cite{Chen2023,Qian2023,Deng2023,Tang2023,Shi2023} further improve the efficiency and 3D consistency.
Our OctFusion can generate 3D shapes in a feed-forward manner within several seconds.

\subsection{Unified Diffusion Models}
To enable diffusion models to generate high-resolution images, a commonly-used strategy is to train cascaded diffusion models~\cite{Ho2022a,Nichol2021,Saharia2022,Ramesh2022}.
To reduce the complexity of the training process and share weights and computation across different levels of resolutions, several approaches~\cite{Emiel2023,Jabri2022,Chen2023a,Gu2023} propose to train a unified diffusion model to directly generate high-resolution images.
UniDiffuser~\cite{Bao2022} proposes a unified diffusion model to model the joint distribution of multi-modal data, which can generate diverse and high-quality samples with different modalities.
The design of OctFusion is inspired by these pioneering works which training an unified U-Net for different resolutions. \looseness=-1
\section{Method} \label{sec:method}

\begin{figure*}[ht]
  \centering
  \includegraphics[width=\linewidth]{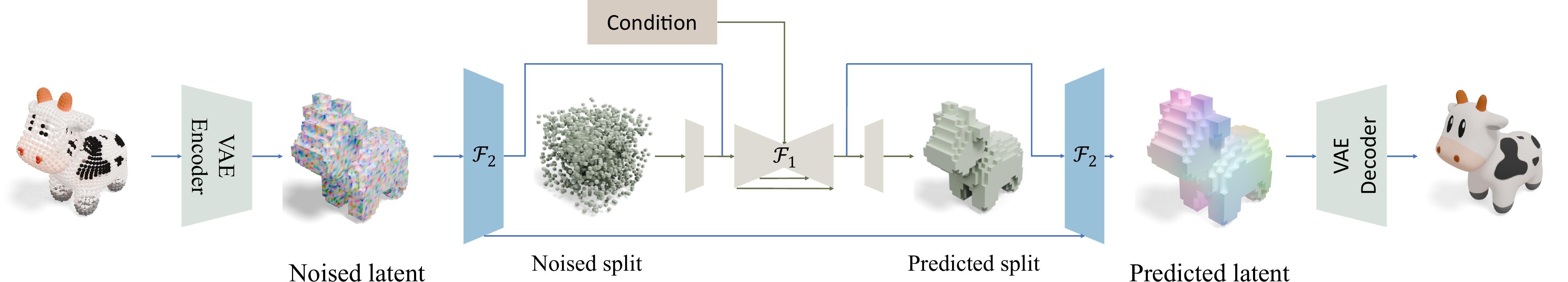}
  \caption{Overview.
  Given a point cloud, possibly with color information, an octree is firstly constructed.
  Then, a VAE is trained to learn latent features on all octree leaf nodes, which can be decoded to continuous distance fields and color fields.
  Next, the first stage model $\mathcal{F}_1$ is trained to predict splitting signals from its noised version, presented as the noised and predicted octree in the figure. Image or text conditions can be optionally provided to guide the generation.
  Then second stage model $\mathcal{F}_2$ predicts latent features, presented as the colors on octree leaf nodes.
  Notably, the weights of U-Net are shared across different octree levels, which plays an important role in saving computation and memory as well as improving performance.
  }
  \label{fig:octfusion}
\end{figure*}

\begin{figure}[t]
  \centering
  \includegraphics[width=\linewidth]{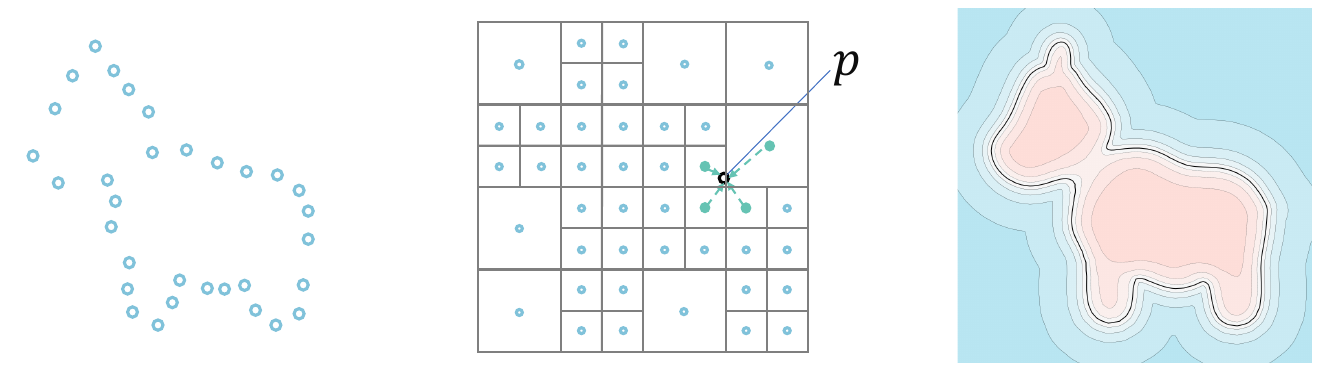}
  \caption{Octree-based Latent Representation. Here 2D figures are used for better visualization.
  Left: An input point cloud sampled from a 3D shape.
  Middle: The octree structure constructed from the point cloud and latent features for octree leaf nodes produced by the VAE. The latent features are shown as solid dots.
  Right: The continuous SDF field reconstructed decoded from the latent features. The field value of an arbitrary query point $p$ is computed with the MPU module.
   }
  \label{fig:representation}
\end{figure}

\subsection{Overview}
Our goal is to efficiently generate high-resolution 3D shapes with diffusion models.
The inherent dilemma is the trade-off between the resolution of 3D shapes and the efficiency of the diffusion model.
To address this challenge, we propose octree-based latent representations and a unified diffusion model for the efficient generation of continuous shapes with resolution of up to $1024^3$.
The overview of our method is shown in Fig.~\ref{fig:octfusion}.
Specifically, we first train a variational autoencoder (VAE) to learn octree-based latent representations for 3D shapes, which can be decoded into continuous signed distance fields (SDFs).
Then, we train an octree-based diffusion model to generate the octree structures and latent features.
We next elaborate on the octree-based latent representation and the diffusion model in Section~\ref{sec:representation} and Section~\ref{sec:diffusion}, respectively.

\subsection{Octree-based Latent Representation} \label{sec:representation}

\subsubsection{Shape Representation}
We encode 3D shapes with octree-based latent representations, which can be decoded into continuous fields, such as signed distance fields (SDFs).
Given a mesh or a point cloud, we convert it to an octree by recursive subdividing nonempty octree nodes until the maximum depth is reached.
All leaf nodes of the octree form an adaptive partition of the 3D volume.
Inspired by~\cite{Wang2022,Ohtake2003}, we store latent features on each leaf node, which are decoded to local SDFs by a shared MLP.
Then, we blend local SDFs into global continuous fields using a multi-level partition-of-unity (MPU) module.
A 2D illustration of the proposed representation is shown in Fig.~\ref{fig:representation}.

For an octree, denote the $i^{th}$ leaf node as $v_i$, with its center as $o_i$, its cell size as $r_i$, and the associated latent feature as $f_i$.
We compute the SDF of an arbitrary query point $p$ using MPU as follows:
\begin{align}
    F_{sdf}(p) &= \frac{\sum_i w_i(x) \cdot \Phi_{sdf}(x, f_i)}
                      {\sum_i w_i(x)},
    \label{equ:mpu}
\end{align}
where $x = (p - o_i) / r_i$, representing the local coordinates of $p$ relative to $o_i$,
$w_i(x)$ is a locally-supported linear B-Spline function,
and $\Phi_{sdf}(x, f_i)$ is a shared MLP that maps the local coordinates $x$ and the latent feature $f_i$ to the SDF value at $p$.
Since $w_i(x)$ and $\Phi_{sdf}(x, f_i)$ are all continuous functions, $F_{sdf}(x)$ are guaranteed to be continuous~\cite{Ohtake2003,Wang2022}.
$\Phi_{sdf}(x, f_i)$ is fully differentiable, and its evaluation is efficient since the weight function $w_i(x)$ is locally supported.

\subsubsection{Octree-based Variational Autoencoders}
We use the variational autoencoder (VAE)~\cite{Kingma2013} built upon dual octree graph networks~\cite{Wang2022} to learn the octree-based latent representation.
The encoder of the VAE takes an octree built from a point cloud as input and outputs a latent feature for each leaf node;
the decoder reconstructs the continuous SDF from the latent features.

We precompute the ground-truth SDF for each shape in the training set.
To train the VAE, we sample a set of points $\mathcal{Q}$ uniformly in the 3D volume and minimize the following loss function to reconstruct the SDF:
{\small
\begin{equation}
  L_{sdf} = \frac{1}{N_\mathcal{Q}}  \sum_{x \in \mathcal{Q}} \left( \lambda_s \| F_{sdf}(x) - D(x)  \|_2^2 + \| \nabla F_{sdf}(x) -  \nabla D(x) \|_2^2 \right),
  \label{equ:sdf}
\end{equation}
}
where $D(x)$ and $\nabla D(x)$ are the ground-truth SDF and the corresponding gradient at the sampled point $x$, respectively, and $\lambda_s$ is set to 200.
The second term in Eq.~\ref{equ:sdf} is used to encourage the predicted SDF to be smooth~\cite{Wang2022}.
To improve the efficiency of the following diffusion model, we also reduce the depth of the original octree with the VAE encoder.
Thus, there is an additional binary cross-entropy loss $L_{octree}$ for the splitting status of each octree node in the decoder following~\cite{Wang2022}.
Finally, we use a KL-divergence loss $L_{KL}$ to regularize the distribution of latent features~\cite{Kingma2013}  to be similar to a standard Gaussian distribution.

In summary, the loss function for the VAE is
\begin{equation}
    L_{VAE} = L_{sdf} + L_{octree} + \lambda L_{KL},
    \label{equ:vae}
\end{equation}
where $\lambda$ is set to 0.1 to balance the affect of $L_{KL}$.
The network of VAE comprises residual blocks built upon dual octree graph networks~\cite{Wang2022}, downsampling, and upsampling modules, which are detailed in the supplementary material.

\subsection{Octree-Based Diffusion Model} \label{sec:diffusion}

\subsubsection{Diffusion Models}
A denoising diffusion model~\cite{Kingma2021,Ho2020} consists of a forward and a reverse process.
The forward process is a fixed Markov chain that transforms the data distribution to a Gaussian distribution $\mathcal{N}(0,\boldsymbol{I})$ by iteratively adding noise with the following formula:
\begin{equation}
    x_t = \sqrt{\alpha(t)}x_0 + \sqrt{1-\alpha(t)}\epsilon,
\label{equ:diffusion_noise}
\end{equation}
where $x_0$ is an input sample, $\epsilon$ is a unit Gaussian noise, $t$ is a uniform random time step in $[0,1]$, and $\alpha(t)$ is a monotonically decreasing function from 0 to 1.
The reverse process maps the unit Gaussian distribution to the data distribution by removing noise.
The prediction from $x_t$ to $x_0$ is modeled by a neural network $\mathcal{F}(x_t,t)$.
The network is trained with the following denoising loss:
\begin{equation}
    L_{diffusion}(x_0) = {E}_{\epsilon,t} \| \mathcal{F}(x_t,t) - x_0 \|_2^2.
    \label{equ:diffusion_loss}
\end{equation}
After training, we leverage the trained model to sample a generative result from the standard Gaussian distribution~\cite{Ho2020}.

\subsubsection{Octree-based Diffusion Model}
Our octree-based representation is determined by the splitting status and the latent feature of each octree node.
The latent features are continuous signals, and we also regard the splitting status as a 0/1 continuous signal, with $0$ indicating no splitting and $1$ indicating splitting.
Then, we follow Eq.~\ref{equ:diffusion_noise} to add noise to the splitting signal and the latent feature of each octree node to get a noised octree.
The goal of our OctFusion is to revert the noising process by predicting clean signals for all octree nodes from a noised octree.
To this end, we train an octree-based U-Net~\cite{Wang2022,Wang2017} by minimizing the loss function defined in Eq.\ref{equ:diffusion_loss}.
The predicted splitting signals are rounded to 0/1 and used to generate the octree structure, and the predicted latent features are used to reconstruct the continuous fields. \looseness=-1

We force the octree to be full when the depth is less than $4$ and denote its maximum depth as $D$.
The noise is added to octree nodes from depth $4$ to $D$; thus, we need to denoise for multiple octree levels.
Instead of training a separate U-Net for each octree level in the spirit of cascaded diffusion models~\cite{Zheng2023,Ren2024}, we propose a unified U-Net that consists of multiple stages and can take noise signals at different octree levels as input and predict the corresponding clean signals.
The detailed architecture of our OctFusion is shown in Fig.~\ref{fig:octfusion}, where each stage of the U-Net is marked with a different color and is responsible for processing octree nodes at a specific depth.
We denote the $i^{th}$ stage of the U-Net as $\mathcal{F}_i$.
The first stage $\mathcal{F}_1$ processes octree nodes with depth $4$, which are equivalent to full voxels with resolution $16^3$ since the octree is full at this depth.
We store an $8$-channel 0/1 signal for each octree node at depth $4$, with which the octree nodes can be split to depth $6$.
Specifically, if all the 0/1 signals in 8 channels are 0, the current octree node will not be split. If at least one channel is 1, the current octree node will be split to 8 child nodes to obtain the next level octree with depth 5. Then, the 0/1 signals of the 8 channels will correspond to the newly split 8 child nodes. If the signal is 0, the corresponding child node will not be split and if it is 1, the corresponding child node will be further split. Therefore, we can finally obtain the octree with depth 6.
Sequentially, $\mathcal{F}_2$ processes octree nodes with depth $6$ and generated octree with depth $8$, until the last octree depth $D$ is reached.
The last stage $\mathcal{F}_k$ processes octree nodes with depth $D$ and predicts clean latent features for all leaf nodes. \looseness=-1

Although all stages of the U-Net can be trained jointly, we empirically find that training the U-Net stage by stage can produce quantitatively better results.
Specifically, we first train $\mathcal{F}_1$ at depth $4$, then fix the weights of $\mathcal{F}_1$ and train $\mathcal{F}_2$, until all stages of the U-Net are is trained.
Notebly, when training $\mathcal{F}_i$, we reuse the trained weights of $\mathcal{F}_{i-1}$, which enables parameter sharing across different octree levels and is beneficial for training with limited data.
It also simplifies the network architecture and improves the efficiency of the model by avoiding the training of multiple separate U-Nets from scratch, greatly reducing the training cost when the octree depth is large.
The U-Net uses similar modules as VAE and uses additional attention modules following~\cite{Zheng2023}.
The detailed network architecture is provided in the supplementary material.

\begin{figure}[t]
  \centering
  \includegraphics[width=\linewidth]{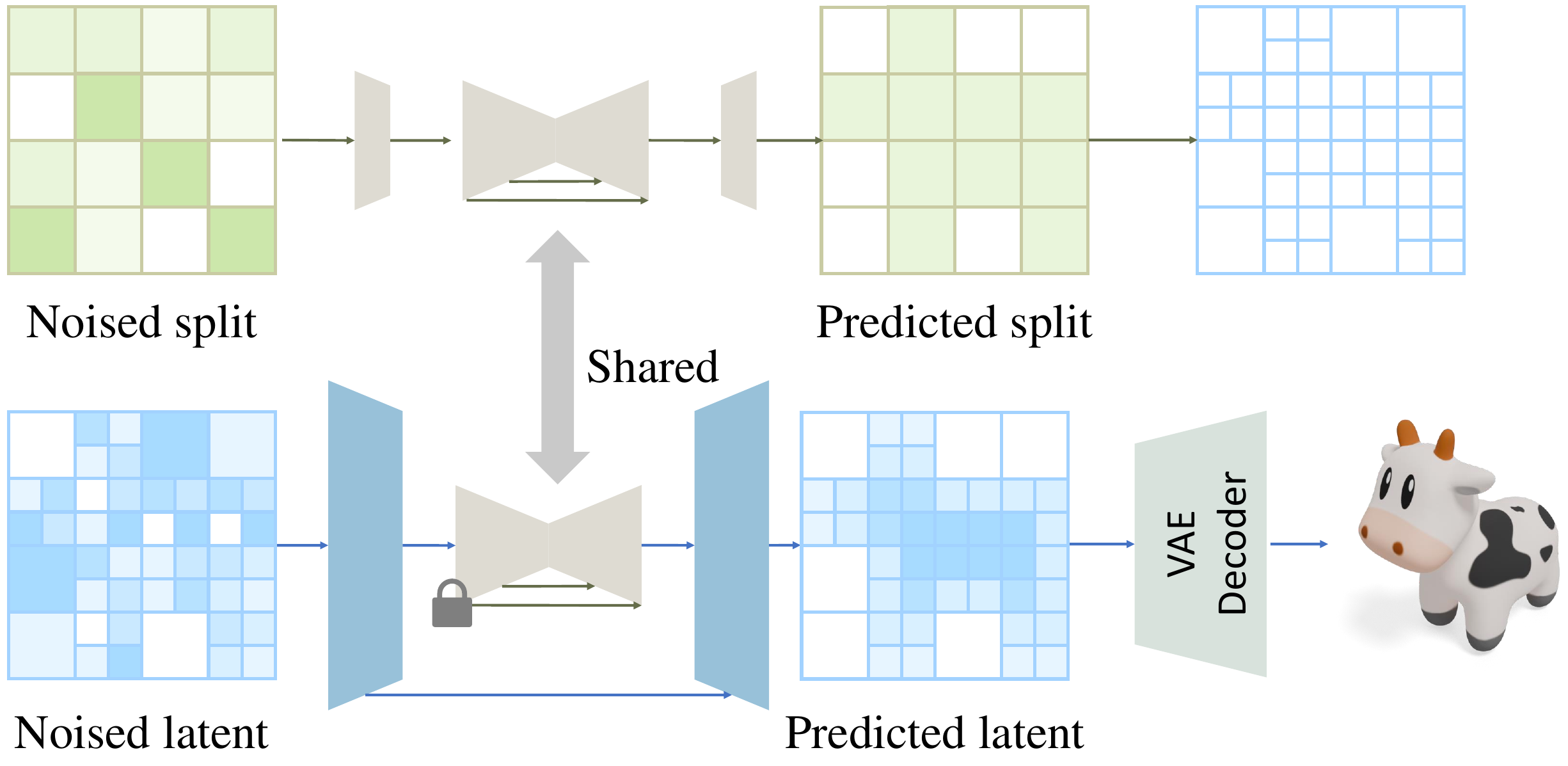}
  \caption{Inference of OctFusion.
  Here 2D figures are used for better visualization.
  Top: The first stage model $\mathcal{F}_1$ on full octree with depth 4 is used to generate the splitting signals.
  Bottom: A dual octree graph convolution network $\mathcal{F}_k$ is used to generate latent features for all leaf nodes, which will be further decoded to the continuous fields.
  }
  \label{fig:sample}
\end{figure}

To generate a result, we first sample random noise at depth $4$ and generate the splitting signals with $\mathcal{F}_1$, with which the octree is grown to depth $6$; then we generate the splitting signals at depth $6$ and grow the octree to depth 8 with $\mathcal{F}_2$, and so on.
In the last stage, we generate latent features for all leaf nodes with $\mathcal{F}_k$, which are further decoded to continuous SDFs with the decoder of VAE. A 2D illustration of the sample pipeline is shown in Fig.~\ref{fig:sample}.
We then extract the zero-level set of SDFs as generated meshes with Marching Cubes~\cite{Lorensen1987}.

\subsubsection{Textured Shape Generation}
Our method can be easily extended to generate color fields.
Specifically, we append another latent feature $c_i$ on octree leaf node $v_i$, which can be decoded into color fields with another shared MLP $\Phi_{color}(x, c_i)$ in a similar manner as Eq.~\ref{equ:mpu}.
And we can learn latent features for color fields with the VAE, using a similar regression loss for color reconstruction.
After training the diffusion model for the SDF fields, we can train another diffusion model for the color fields with the same architecture.
With the color fields, we can assign an RGB color to each vertex of the generated mesh.
We also verify the effectiveness of our method for generating textured shapes in the experiments.

\section{Experiments} \label{sec:experiments}

In this section, we verify the effectiveness and generative quality of OctFusion from a variety of respects.

\subsection{3D Shape Generation} \label{model_training}

\subsubsection{Dataset}
We choose 5 categories from ShapeNetV1~\cite{Chang2015} following LAS-Diffusion~\cite{Zheng2023} and use the same training, evaluation, and testing data split to train our model and conduct comparisons.
The categories include \texttt{chair}, \texttt{table}, \texttt{airplane}, \texttt{car}, and \texttt{rifle}.
Some meshes in ShapeNet are non-watertight and non-manifold; we repair them following~\cite{Wang2022} and normalize them to the unit cube.
We further convert the repaired meshes to signed distance fields (SDFs) for the VAE's training. \looseness=-1

\subsubsection{Training Details}
To train the VAE, we sample $200k$ points with oriented normals on each repaired mesh and build an octree with depth 8 (resolution $256^3$).
In each training iteration, we randomly sample $50k$ SDF values from the 3D volume for each shape to evaluate the loss function~\ref{equ:vae}.
We train the VAE with the AdamW~\cite{Loshchilov2017} optimizer for 200 epochs with batch size 8 on 2 Nvidia A40 (48G) GPUs.
The initial learning rate is set as $10^{-3}$ and decreases to $10^{-5}$ linearly throughout the training process.
The encoder of VAE downsamples the input octree to depth 6 (resolution $64^3$) and outputs a 3-dimensional latent code per octree leaf node.
We also train our diffusion models with the AdamW optimizer.
The U-Net of OctFusion contains 2 stages, which are trained for 4000 epochs in less than one day and 500 epochs for two days on 4 Nvidia 4090 GPUs, respectively, with a fixed learning rate of $10^{-4}$.
To compare with other methods,  we train our geometry-only OctFusion model with both unconditional and category-conditional settings.
For the unconditional generation, we train our OctFusion on each category.
For category-conditional generation, we train our OctFusion on 5 categories data with the label embedding as conditional input.

\begin{figure*}[t!]
  \includegraphics[width=0.98\textwidth]{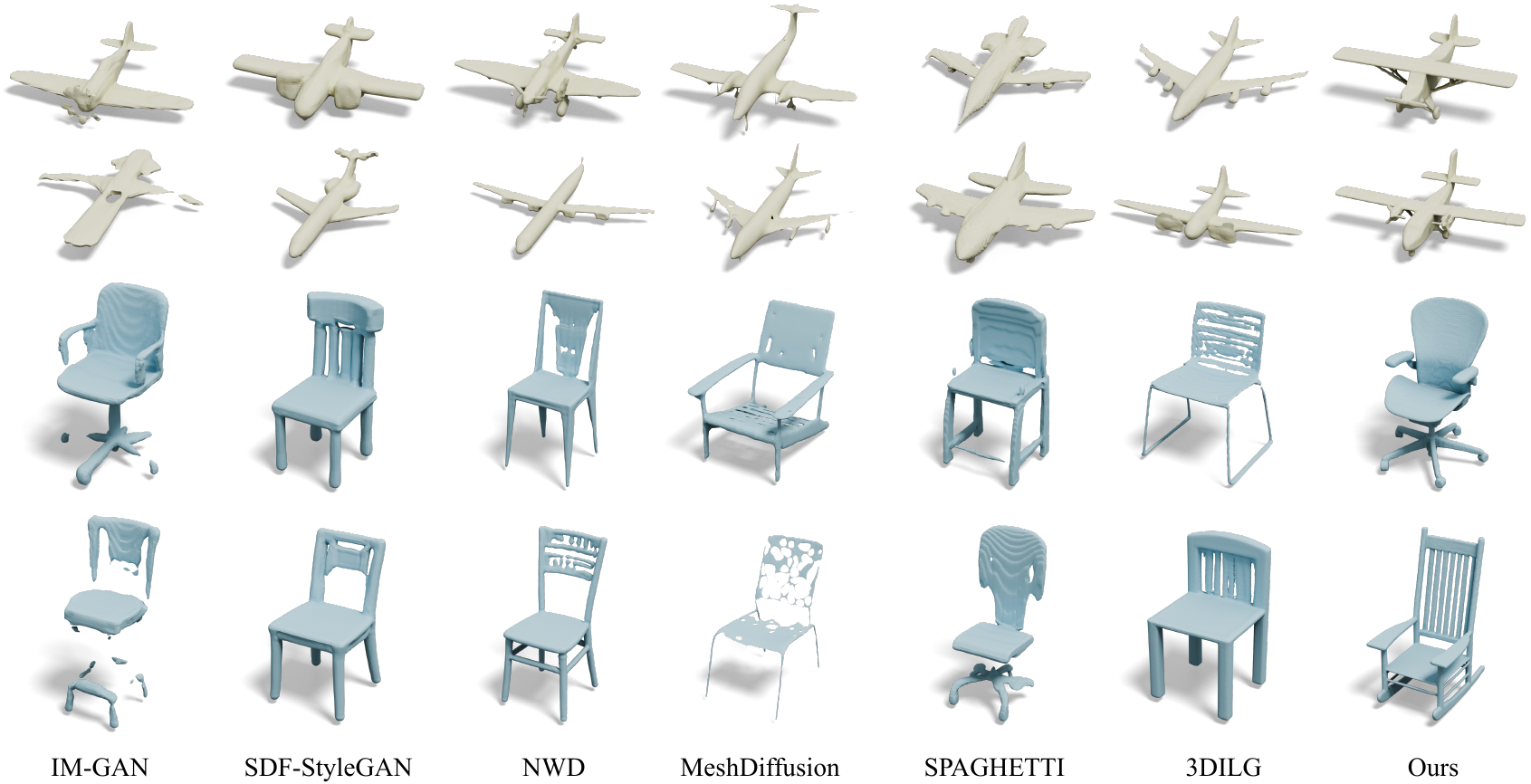}
  \caption{Examples of generated airplanes and chairs obtained by OctFusion and other state-of-the-art generation models.}
  \label{fig:diffusion_compare}
\end{figure*}

\subsubsection{Evaluation Metrics}
In line with prior research~\cite{Zheng2023,Zhang2023a}, we utilize the \emph{shading-image-based FID} as the primary metric to assess the quality and diversity of the generated 3D shapes.
Specifically, we render the generated meshes from 20 uniformly distributed viewpoints, and these images are used to calculate the FID scores against the rendered images from the original training dataset.
A lower FID score indicates better generation quality and diversity.
Additionally, we adopt the COV, MMD, and 1-NNA metrics~\cite{Hui2022, Achlioptas2018, Yang2019}, in which COV measures the coverage of the generated shapes, MMD evaluates the fidelity of the generated shapes, and 1-NNA assesses the diversity of the generated shapes.
For these metrics, we generate 2000 shapes and compare them with the test set  following SDF-StyleGAN~\cite{Zheng2022} and LAS-Diffusion~\cite{Zheng2023}.
For each mesh, we uniformly sample 2048 points and normalize them within a unit cube to compute the Chamfer distance (CD) and Earth mover's distance (EMD).
Lower MMD, higher COV, and a 1-NNA value closer to 50\% indicate better quality.

\begin{table}[!t]
    \caption{The quantitative comparison of \emph{shading-image-based FID} between results generated by OctFusion and other methods. The superscript $^\dagger$ and $^\ddagger$ denote unconditional and category-conditional version of corresponding methods, respectively. Note that Wavelet-Diffusion was trained on 3 categories only and SPAGHETTI was trained on 2 categories only.}
    \label{tab:comparison}
    \tablestyle{6pt}{1.2}
    \begin{tabular}{lccccc}
      \toprule
      \textbf{Method} & \textbf{Chair} & \textbf{Airplane} & \textbf{Car} & \textbf{Table} &  \textbf{Rifle} \\
      \midrule
      IM-GAN & 63.42 & 74.57 & 141.2 & 51.70 & 103.3 \\
      SDF-StyleGAN & 36.48 & 65.77 & 97.99 & 39.03 & 64.86 \\
      Wavelet-Diffusion & 28.64 & 35.05 & N/A & 30.27 & N/A \\
      MeshDiffusion & 49.01 & 97.81 & 156.21 & 49.71 & 87.96 \\
      SPAGHETTI & 65.26 & 59.21 & N/A & N/A & N/A \\
      LAS-Diffusion$^\dagger$ & 20.45 & 32.71 & 80.55 & 17.25 & 44.93 \\
      XCube & 18.07 & \textbf{19.08} & 80.00 & N/A & N/A \\
      OctFusion$^\dagger$ & \textbf{16.15} & 24.29 & \textbf{78.00} & \textbf{17.19} & \textbf{30.56}  \\
      \midrule
      3DILG & 31.64 & 54.38 & 164.15 & 54.13 & 77.74 \\
      3DShape2VecSet & 21.21 & 46.27 & 110.12 & 25.15 & 54.20 \\
      LAS-Diffusion$^\ddagger$ & 21.55 & 43.08 & 86.34 & \textbf{17.41} & 70.39  \\
      OctFusion$^\ddagger$ & \textbf{19.63} & \textbf{30.92} & \textbf{80.97} & 17.49 & \textbf{28.59} \\
      \bottomrule
    \end{tabular}
\end{table}

\begin{table}[t]
    \caption{Additional quantitative comparison between different models.
    The units of CD and EMD are $10^{-3}$ and $10^{-2}$, respectively.}
    \label{tab:cov_mmd}
    \tablestyle{3.8pt}{1.2}
    \begin{tabular}{lccccccccc}
      \toprule
      \multirow{2}{*}{\textbf{Method}} & \multicolumn{2}{c}{\textbf{COV(\%)$\uparrow$}} && \multicolumn{2}{c}{\textbf{MMD$\downarrow$}} && \multicolumn{2}{c}{\textbf{1-NNA(\%)$\downarrow$}} \\
      \cline{2-3} \cline{5-6} \cline{8-9}
      & \textbf{CD} & \textbf{EMD} & & \textbf{CD} & \textbf{EMD} && \textbf{CD} & \textbf{EMD} \\
      \midrule
      IM-GAN        & \textbf{57.30} & 49.48  && \textbf{13.12} & 17.70 && 62.24 & 69.32 \\
      SDF-StyleGAN  & 52.36 & 48.89 && 14.97 & 18.10 && 65.38 & 69.06 \\
      Wavelet-Diffusion & 52.88 & 47.64       && 13.37 & \textbf{17.33} && \textbf{61.14} & 66.92 \\
      LAS-Diffusion & 53.76 & 52.43 && 13.79 & 17.45          && 64.53 & 65.15 \\
      OctFusion  & 53.59 & \textbf{53.17} && 13.78 & 17.44     && 63.02 & \textbf{63.72} \\
      \bottomrule
    \end{tabular}
\end{table}

\subsubsection{Comparisons}
We conduct comparisons with representative state-of-the-art generative models, including IM-GAN \cite{Chen2019}, SDF-StyleGAN \cite{Zheng2022}, Wavelet-Diffusion \cite{Hui2022}, 3DILG \cite{Zhang2022a}, MeshDiffusion \cite{Liu2023c}, SPAGHETTI \cite{Hertz2022}, LAS-Diffusion \cite{Zheng2023}, 3DShape2VecSet \cite{Zhang2023a} and XCube \cite{Ren2024}.
Among these methods, IM-GAN and SDF-StyleGAN are GAN-based methods, and the others are diffusion-based methods;
3DILG and 3DShape2VecSet are category conditional models; the others are unconditional generative models trained on each category separately. It is worth noting that we \emph{did not conduct any post-processing} on our generated meshes extracted by marching cube algorithm~\cite{Lorensen1987}, while some other methods such as MeshDiffusion~\cite{Liu2023c} removes isolated meshes of tiny sizes and applies both remeshing and the standard Laplace smoothing on all the generated meshes.

\begin{figure*}[t!]
  \centering
  \includegraphics[width=0.98\textwidth]{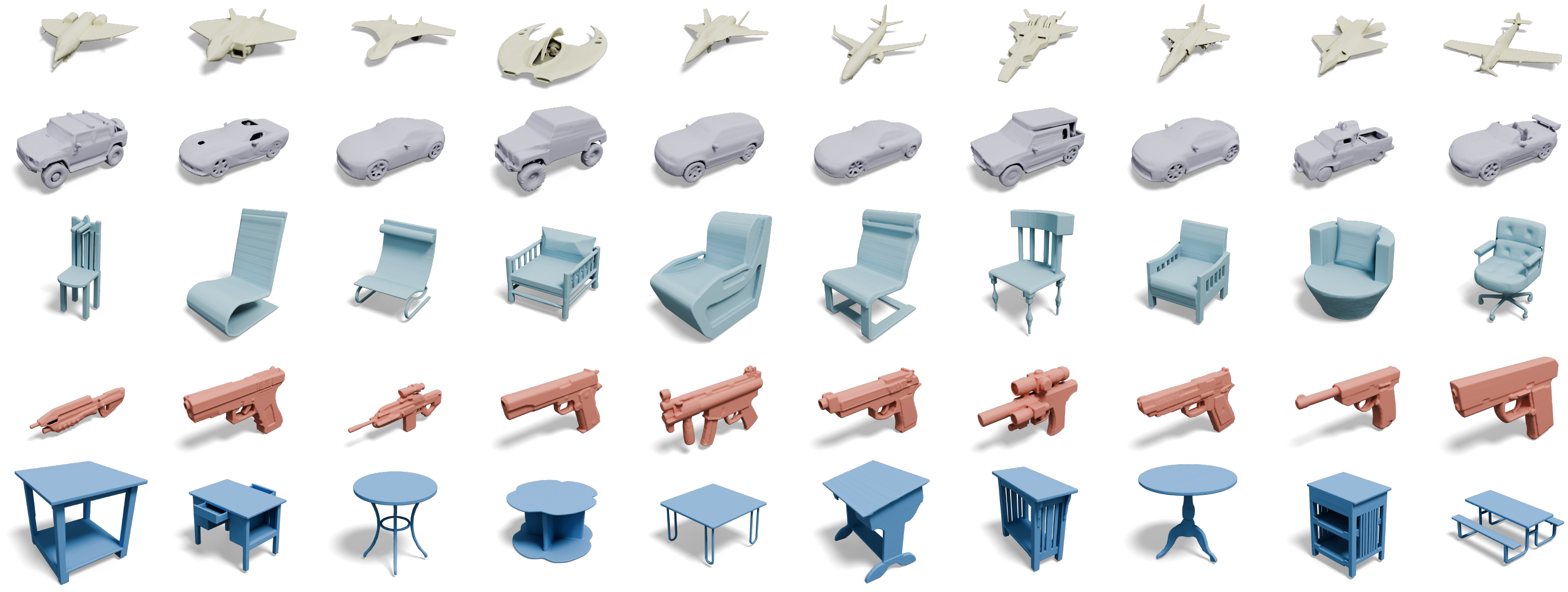}
  \caption{Unconditional generation results.
  Please zoom in for better inspection of the complex geometry and thin structure.}
  \label{fig:uncond}
\end{figure*}

\begin{figure}[t]
  \includegraphics[width=\columnwidth]{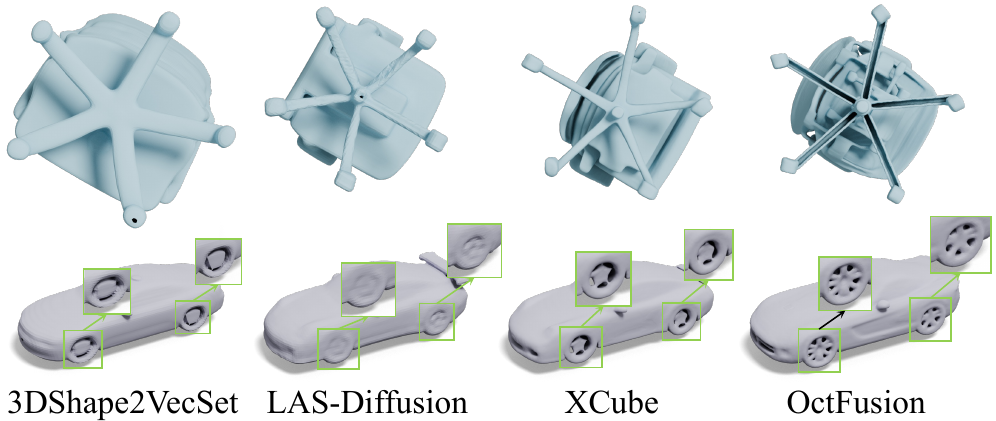}
  \caption{Detailed comparison of OctFusion with 3DShape2VecSet, LAS-Diffusion and XCube on chair and car category. From which we can see that our method has a greater capability of capturing geometry details such as the fluting of swivel chairs and the wheel hubs of cars.
  }
  \label{fig:detail_compare}
\end{figure}

Table~\ref{tab:comparison} reports the comparison of the \emph{shading-image-based FID}.
OctFusion achieves the best-generating quality on average under both unconditional and category-conditional settings compared to all previous methods, demonstrating its superiority.
The margin of improvement is more significant in the categories of \texttt{chair} and \texttt{rifle}, which contain more complex structures and thin details.
The comparison with 3DILG, 3DShape2VecSet, XCube is for reference only, as their training data are not exactly the same as those of the other methods.
Table~\ref{tab:cov_mmd} reports the quantitative comparison results of COV, MMD, and 1-NNA metrics with models using the same train/eval/test split as ours because these metrics require to be calculated and compared with test set. From Table~\ref{tab:cov_mmd} we can see that our method achieves the best COV(EMD) and 1-NNA(EMD) among all the competing methods and is comparable with Wavelet-Diffusion~\cite{Hui2022} on MMD. However, the drawbacks of COV, MMD, and 1-NNA have been confirmed by previous work~\cite{Zheng2022}. Thus, we primarily focus on comparing \emph{shading-image-based FID}. \looseness=-1

\begin{figure}[h]
  \includegraphics[width=0.98\columnwidth]{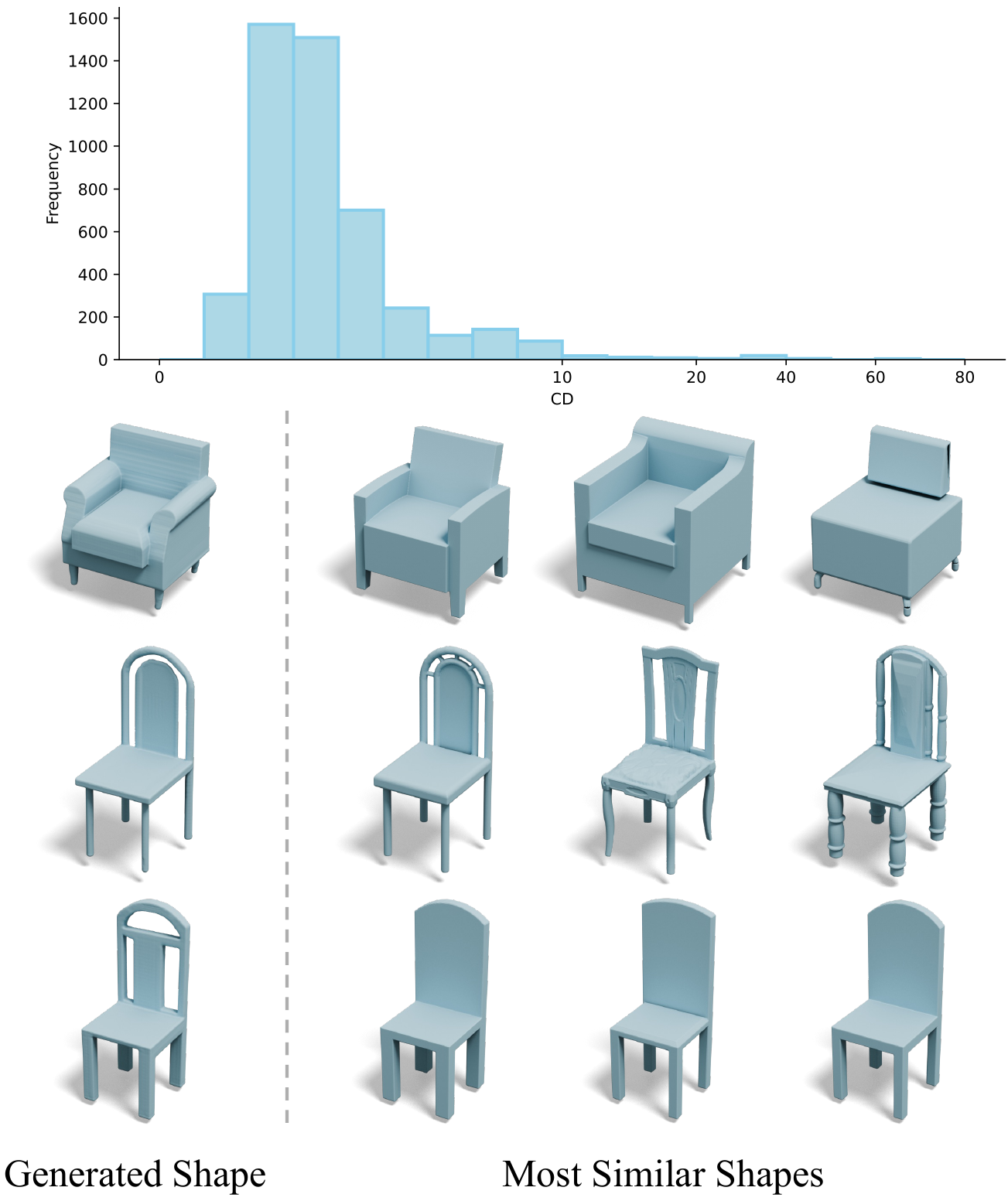}
  \caption{\textbf{Top:} The histogram on the distribution of Chamfer distances between the generated chairs and the most similar one in training set. The unit of CD is $10^{-3}$. \textbf{Bottom:} The generated shapes (Left) and the 3 nearest shapes (right) retrieved from the training dataset according to their Chamfer distances.}
  \label{fig:diversity}
\end{figure}

Fig.~\ref{fig:diffusion_compare} provides qualitative results of airplanes and chairs generated by different methods.
We can see that the results IM-GAN, SDF-StyleGAN, and  MeshDiffusion contain severe distortions and artifacts.
Although LAS-Diffusion, 3DShape2VecSet make significant progress in geometry quality, they fail to capture high-detailed geometric features due to the limitation of resolution of their 3D representation, such as the propellers of airplanes and the thin slats of chairs.
Besides Fig.~\ref{fig:diffusion_compare}, we also provide a far more comprehensive qualitative comparison with LAS-Diffusion, 3DShape2VecSet and a more recently proposed method: XCube in Fig.~\ref{fig:detail_compare}. From which we can see that similar to LAS-Diffusion, XCube uses hierarchical sparse voxel and reaches the resolution of $512^3$ on ShapeNet. However, there are still certain difficulties for XCube in modeling the fluting of swivel chairs and the wheel hubs of cars.
On the contrary, our OctFusion generates implicit features on each octree node which demonstrates superiority in capturing the fine geometry details of 3D shapes.
Fig.~\ref{fig:uncond} shows more high-quality and diverse generative results by our unconditional OctFusion model trained on five categories of ShapeNet separately.
And we provide more generative results in the supplementary materials.

\subsubsection{Shape Diversity}
We evaluate the diversity of generated shapes to identify whether OctFusion just simply memorizes the training data.
We conducted this analysis on \texttt{chair} category by retrieving the most similar one to the given generated 3D shape in the training set by Chamfer distance metric.
Fig.~\ref{fig:diversity}-top shows the histogram whose x-axis is Chamer distrance ($\times 10^{-3}$) which demonstrates that most generated shapes are difference from the training set. Fig.~\ref{fig:diversity}-bottom presents some generated samples associated with the three most similar shapes retrieved from the training set.
Comparing our generated shapes with the most similar one, it can be clearly observed that our OctFusion is able to generate novel shapes with high diversity instead of just memorizing all cases in training data.

\subsubsection{Model generalizability}
\begin{figure}[t]
  \includegraphics[width=\columnwidth]{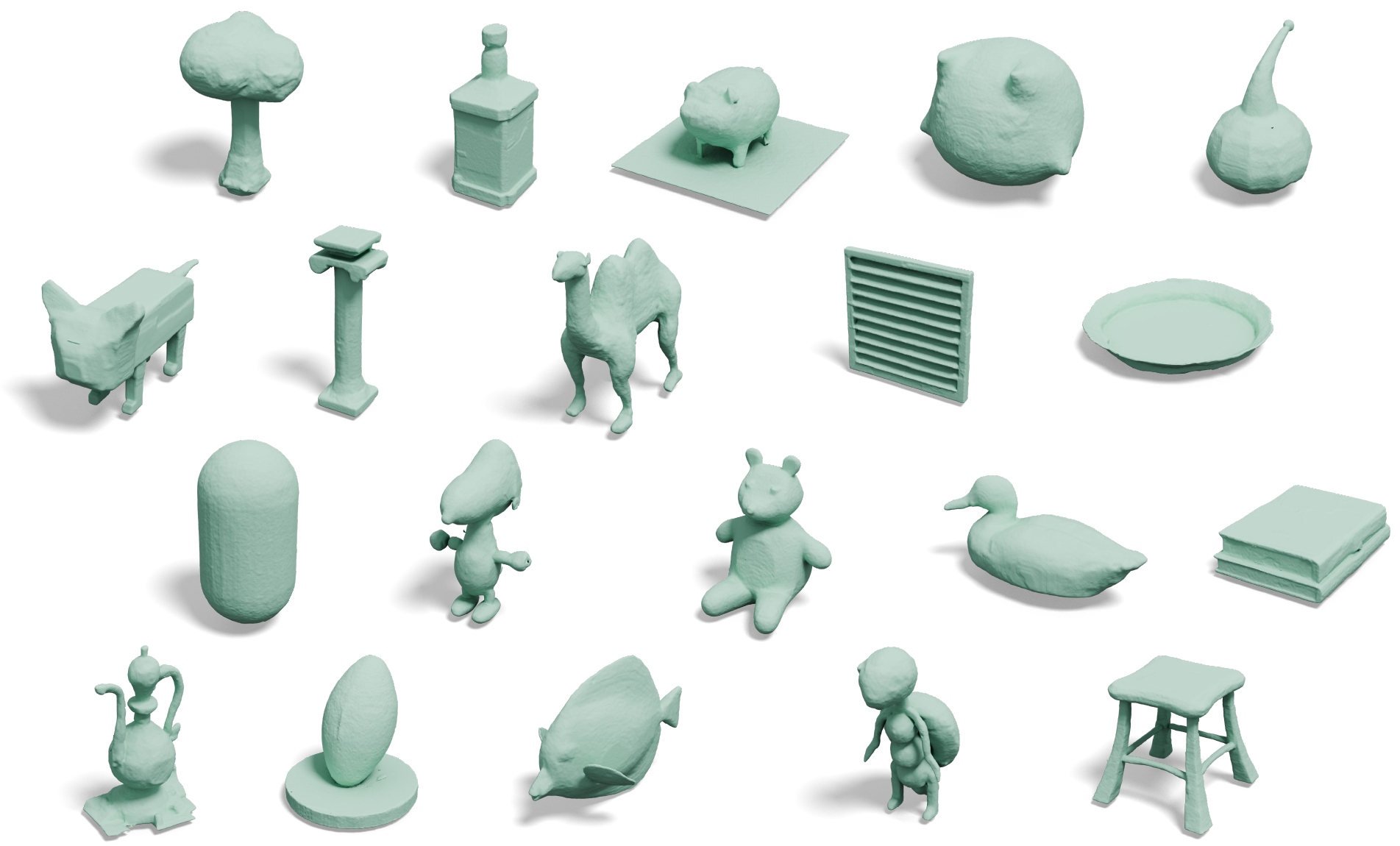}
  \caption{Unconditional generation results on the Objaverse dataset.}
  \label{fig:objaverse}
\end{figure}

We also evaluate the model generalizability of our OctFusion on Objaverse~\cite{Deitke2023}, a recently proposed 3D shape dataset that contains much richer and more diverse 3D objects. For simplicity, we select a subset containing about 10k high-quality meshes in Objaverse provided by LGM\cite{tang2024lgm} and train our OctFusion model with depth 10 (resolution $1024^3$). Fig.~\ref{fig:objaverse} provides unconditional generative results on Objaverse. As we can see from Fig~\ref{fig:objaverse}, although 3D objects in Objaverse exhibit far stronger diversity than a single category of ShapeNet, our method is still capable of generating plausible 3D shapes satisfying this complex data distribution.

\subsubsection{Textured Shape Generation}
Our OctFusion is capable of generating textured 3D shapes by extending the diffusion model to generate additional color latent features.
The color latent features can be trained by attaching another decoder to the VAE while sharing the encoder with the geometry latent features.
We train OctFusion on the same 5 categories from ShapeNet and conduct unconditional generation experiments and evaluations.
The details of dataset preprocessing and training are provided in the supplementary materials. \looseness=-1

\begin{figure}[t]
  \includegraphics[width=\columnwidth]{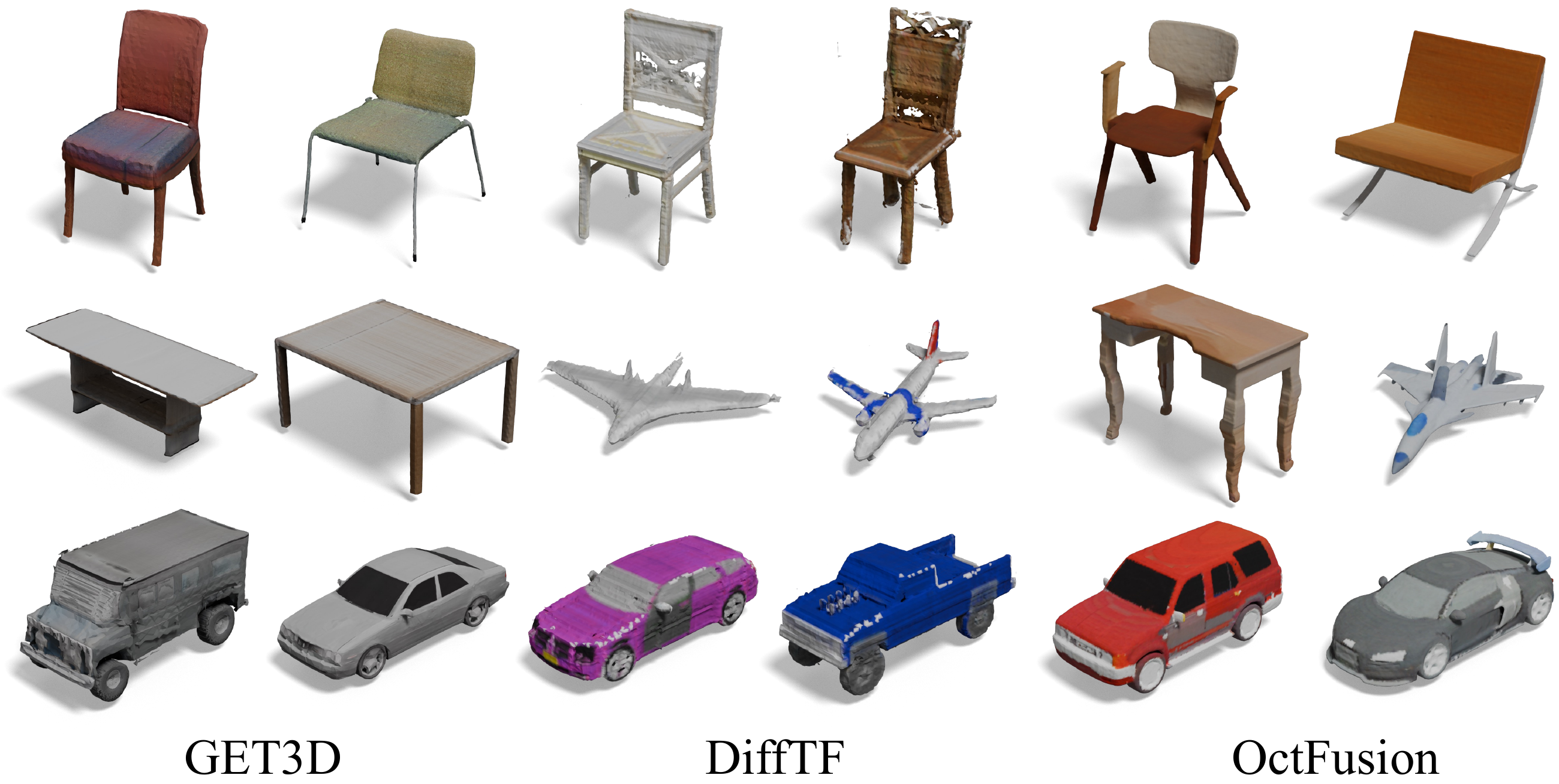}
  \caption{The comparison of generated textured mesh of our method and other state-of-the-arts.}
  \label{fig:color_uncond}
\end{figure}

We compare OctFusion with two recent methods: GET3D \cite{Gao2022} and DiffTF \cite{Cao2024}.
We adopt the \emph{rendering-image-based FID} to evaluate the quality and diversity of the generated textured meshes.
Each generated textured mesh is rendered from 20 uniformly distributed views to RGB images to compute the FID score.
Table~\ref{tab:color_comparison} and Fig.~\ref{fig:color_uncond} provides quantitative and qualitative comparisons.
Our OctFusion achieves the best FID scores on average, demonstrating the superior capability of generating high-quality and diverse textured 3D shapes.
DiffTF generates NeRF as the output; we convert the generated NeRFs to 3D textured meshes using the code provided by the authors.
The rendering of the extracted meshes is different from the NeRF rendering.
Thus, the FID values of DiffTF are for reference.

\begin{table}[t]
  \caption{The quantitative comparison of \emph{rendering-image-based FID} on 3D textured mesh generation. GET3D and DiffTF were trained on 3 categories only. DiffTF is for reference as it generates NeRF instead of textured mesh.}
  \label{tab:color_comparison}
  \tablestyle{8pt}{1.2}
  \begin{tabular}{lccccc}
    \toprule
    \textbf{Method} & \textbf{Chair} & \textbf{Airplane} & \textbf{Car} & \textbf{Table} &  \textbf{Rifle} \\
    \midrule
    GET3D & 51.79 & N/A & \textbf{60.89} & 59.41 & N/A \\
    DiffTF & 64.58 & 90.48 & 137.96 & N/A & N/A \\
    OctFusion & \textbf{31.81} & \textbf{26.64} & 65.58 & \textbf{43.87} & \textbf{41.20} \\
    \bottomrule
  \end{tabular}
\end{table}

\begin{figure}[t!]
\centering
  \includegraphics[width=0.8\columnwidth]{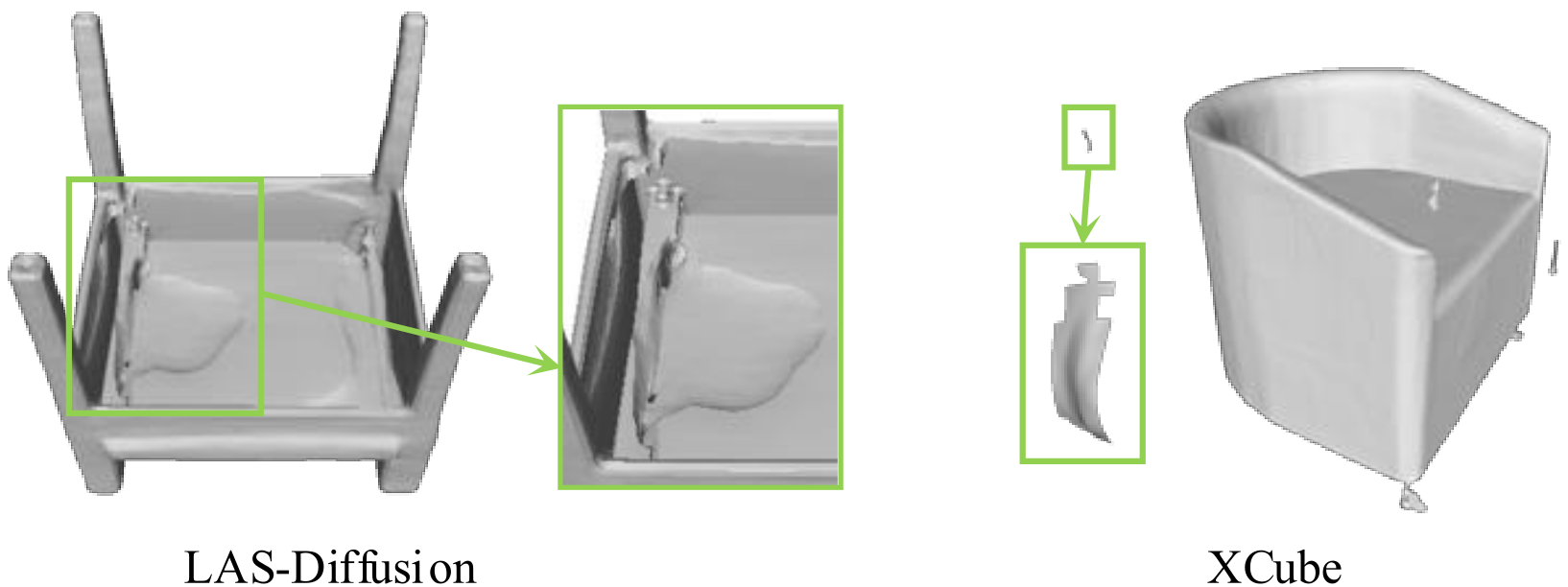}
  \caption{Some failed cases generated by LAS-Diffusion and XCube, which demonstrate the superiority of our method in ensuring the completeness and continuity.
  }
  \label{fig:non-manifold}
  \vspace{-0.2cm}
\end{figure}

\begin{table}[t]
  \tablestyle{2pt}{1.2}
  \caption{Efficiency comparison with LAS-Diffusion and XCube. The average octree node number, the GPU memory and a single forward time cost on a Nvidia 4090 GPU with batch size 1 are reported. The subscript * and ** denote first stage and second stage of corresponding methods, respectively.}
  \begin{tabular}{lcccc}
    \toprule
    \textbf{Method} & \textbf{Node Number*} & \textbf{Node Number**} & \textbf{Memory} & \textbf{Time} \\
    \midrule
    LAS-Diffusion & $262,144$ & $124,156$ & $1.06G$  & $66.1ms$ \\
    XCube & 4096 & 74,761 & $12.76G$  & $135.3ms$ \\
    OctFusion  & 4096 & 11,634 & $0.69G$  & $48.2ms$ \\
    \bottomrule
  \end{tabular}
  \label{tab:efficiancy}
\end{table}

\begin{table}[t!]
  \tablestyle{9pt}{1.2}
  \caption{Training cost comparison with LAS-Diffusion and XCube on the model parameters, number of GPUs, and training time.}
  \begin{tabular}{lccc}
    \toprule
    \textbf{Method} & \textbf{Params} & \textbf{GPUs} & \textbf{Training Time}\\
    \midrule
    LAS-Diffusion & 57M & 8 $\times$ V100 & $\sim$7 days\\
    XCube & 1.6B & 8 $\times$ A100 & $\sim$4 days\\
    OctFusion  & 33M & 4 $\times$ 4090 & $\sim$3 days\\
    \bottomrule
  \end{tabular}
  \vspace{-0.3cm}
  \label{tab:training_cost}
\end{table}

\begin{figure}[t!]
  \includegraphics[width=\columnwidth]{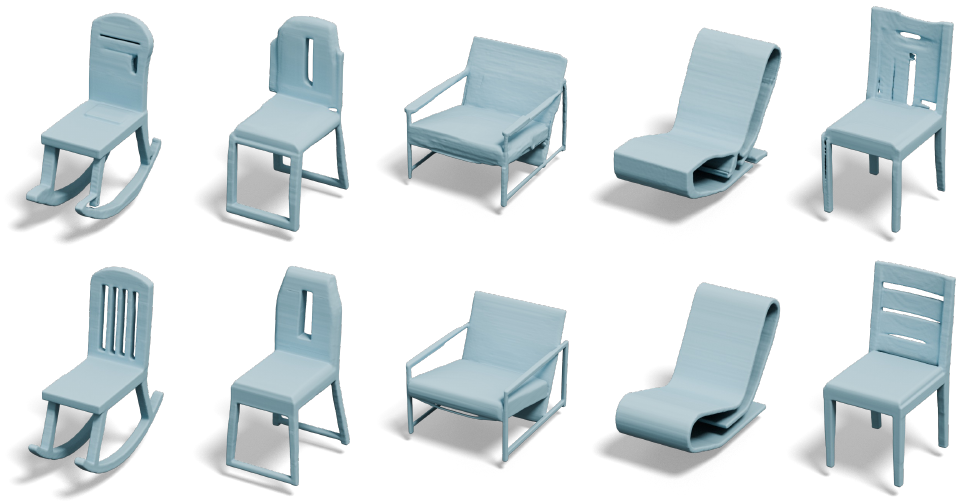}
  \caption{Sharing weights in the U-Net for diffusion. In the 1st row are results without using shared weights, which are worse than results with sharing weights in the 2nd row.
  }
  \label{fig:ablation}
\end{figure}

\subsection{Ablations and Discussions}
In this section, we analyze the impacts of key design choices of OctFusion, including the octree-based latent representation and unified diffusion model.
We choose the category \texttt{chair} to do the ablation studies due to its large variations in structure and topology.

\subsubsection{Octree-based Latent Representation}
\label{sec:ablation}
Here, we discuss the key benefits of our octree-based latent representation over other highly related representations, including plain octrees in LAS-Diffusion~\cite{Zheng2023} and hierarchical sparse voxels in XCube~\cite{Ren2024} and Trellis~\cite{xiang2024structured}.
\begin{itemize}[leftmargin=*,itemsep=2pt]
  \item[-] \textbf{Completeness}.
    The leaf nodes of an octree form a complete coverage of the 3D bounding volume.
    We keep \emph{all} octree leaf nodes in the latent space, which guarantees to contain the whole shape, whereas LAS-Diffusion, XCube and Trellis \emph{prune} voxels and only keep a subvolume, which may lead to holes in the generated shapes, such as the non-manifold meshes shown by the right case in Fig.~\ref{fig:non-manifold}. \looseness=-1
  \item[-] \textbf{Continuity}.
    We merge local implicit fields of all octree leaf nodes to form a global implicit field via the MPU module in Eq.~\ref{equ:mpu}, which is guaranteed to be continuous.
    In contrast, LAS-Diffusion and XCube represent 3D shapes in thick shells with finite discrete voxels and have no such guarantees. As a consequence, they cannot effectively model continuous surfaces which might be truncated due to the presence of the thick shell just like the left case shown in Fig.~\ref{fig:non-manifold}. \looseness=-1
  \item[-] \textbf{Efficiency}.
    We observe that the node number in the octree is significantly reduced compared to LAS-Diffusion and XCube. The reason is that LAS-Diffusion needs to keep a volume shell to cover the shape surface, and XCube keeps all interior voxels of the shape. While we only keep the voxels intersecting with the surface, which makes our network more efficent than LAS-Diffusion and XCube in terms of GPU memory and inference time as shown in Table~\ref{tab:efficiancy}. We also present a quantitative analysis of training costs of different methods in Table~\ref{tab:training_cost}. OctFusion uses the lowest model parameters, GPU memory consumption and training time compared with LAS-Diffusion and XCube, achieving significant resource efficiency.
\end{itemize}

\begin{figure*}[t!]
  \includegraphics[width=0.96\textwidth]{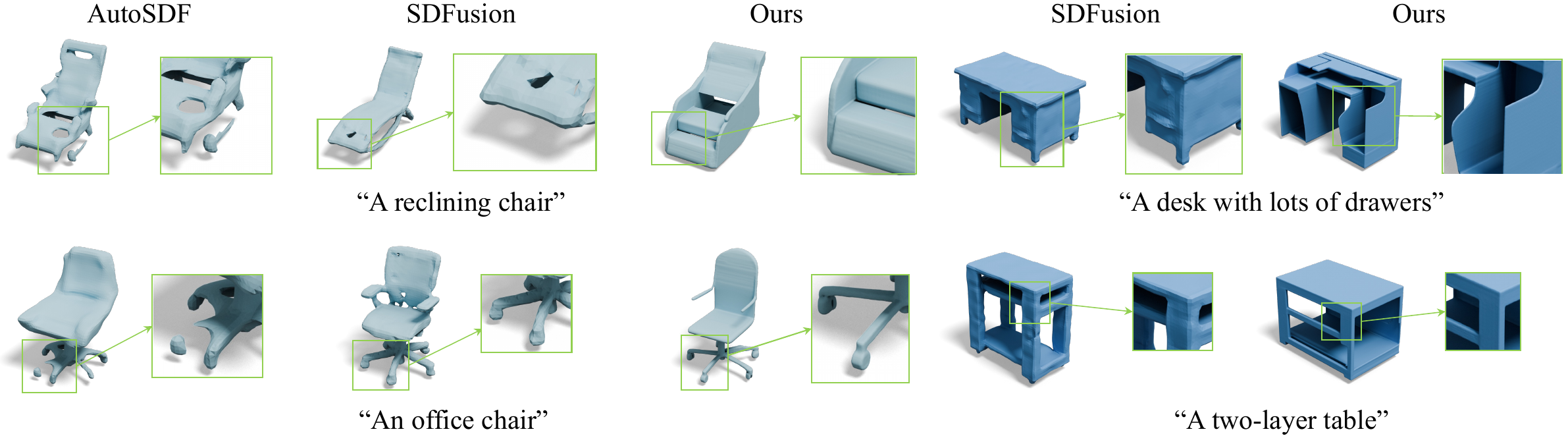}
  \caption{The comparison of our OctFusion with AutoSDF and SDFusion on the Text2Shape dataset. Please zoom
   in for better inspection of the shape quality of our OctFusion and other methods.}
  \label{fig:text}
\end{figure*}

\begin{figure}[t]
  \includegraphics[width=\columnwidth]{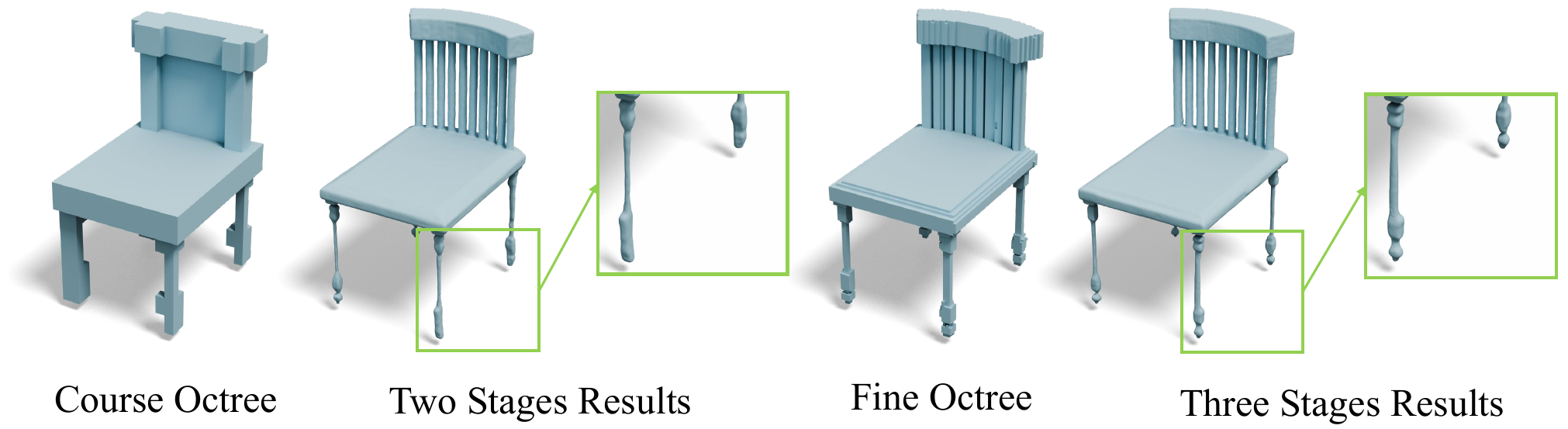}
  \caption{The comparison of OctFusion using more diffusion stages.}
  \label{fig:three_stages}
\end{figure}

\subsubsection{Unified U-Net}
The key insight of our unified U-Net is to share weights across different octree levels.
We conduct a comparison to verify the significance of this design choice by training two variants of OctFusion: V1 without shared weights and V2 with separate weights for different octree levels, which resembles the cascaded diffusion models in LAS-Diffusion and XCube.
The FID of V1 is $22.22$, which is significantly higher than our OctFusion with shared weights, which achieves an FID of $16.15$.
We also visualize generated results of our OctFusion and the variant V1 in Fig.~\ref{fig:ablation}.
Apparently, the results of our OctFusion are more plausible and diverse than V1.
For variant V2, it requires \emph{2 times} more trainable parameters to achieve a comparable FID of $16.00$.
Its time and memory costs are also higher, and its convergence speed is slower than our OctFusion.
Moreover, our OctFusion uses only a single U-Net and greatly reduces the complexity of training and deployment, while the variant V2 requires multiple U-Nets.
This disadvantage will be pronounced when the number of cascades increases. For example, XCube uses up to 3 cascades and needs to train 6 networks in total and its released model has more than 1.6B parameters. On the contrary, our OctFusion has only 33M parameters.

\subsubsection{Deeper Octree}
One key factor to increase the expressiveness of our representation is to increase the depth of the octree.
We conduct an experiment to train OctFusion on a deeper octree with depth 10 (resolution $1024^3$) and train the VAE to extract latent code on an octree with depth 8.
We increase the stages of the U-Net to 3 to match the depth of the generated octree.
Fig.~\ref{fig:three_stages} shows the generative results; it can be seen that OctFusion with three diffusion stages embodies much richer details such as the protruding part of chair legs compared to two stages.
However, the computation cost is greatly increased.
As a result, we choose two stages as our default setting to balance the generative quality and efficiency.

\subsection{Applications}
In this section, we demonstrate applications of OctFusion, including  text/sketch-conditioned generation and  shape texture generation. \looseness=-1

\subsubsection{Text-conditioned generation}
We encode the textual condition with CLIP and then inject extract text features into OctFusion using cross attention.
We use the Text2shape dataset~\cite{Chen2018a,Cheng2023}, that provides textual descriptions for the \texttt{chair} and \texttt{table} categories in ShapeNet.
Fig.~\ref{fig:text} provides qualitative comparisons between our method with AutoSDF~\cite{Mittal2022} and SDFusion~\cite{Cheng2023}.
It can be see that the results of AutoSDF and SDFusion exhibit severe distortions and large amounts of artifacts.
On the contrary, our method is capable of generating shapes with higher quality and delicate structure details such as lots of drawers inside table while conforming to the input textual description. \looseness=-1

\subsubsection{Sketch-conditioned generation}
\begin{figure}[t]
  \includegraphics[width=\columnwidth]{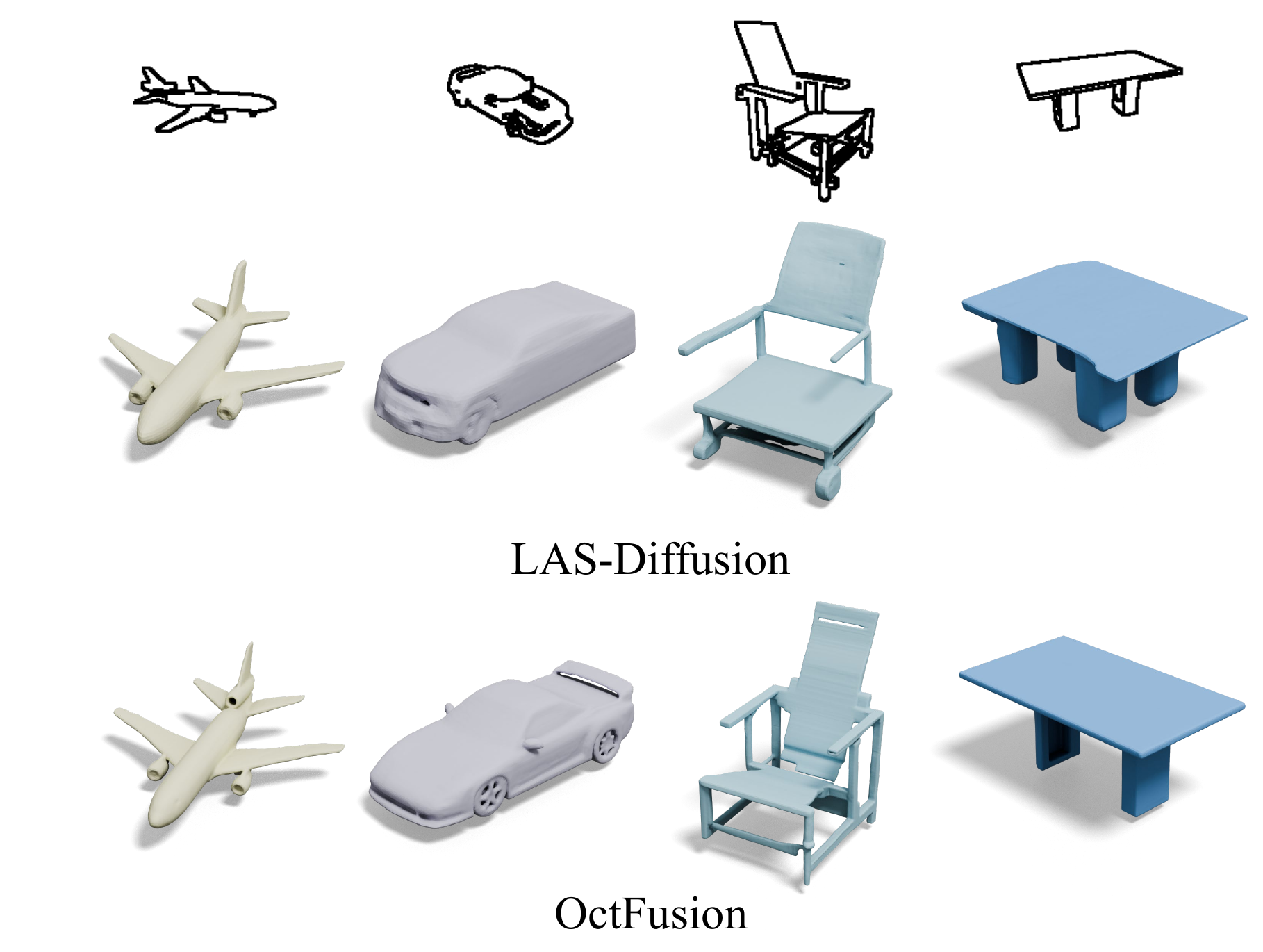}
  \caption{The sketch-conditional generation of OctFusion and LAS-Diffusion.}
  \label{fig:sketch}
\end{figure}

We conduct sketch conditional 3D shape generation using the view-aware local attention proposed in LAS-Diffusion~\cite{Zheng2023} to aggregate sketch information to guide the generation of octrees.
We train our OctFusion with the sketch dataset provided by LAS-Diffusion.
Fig.~\ref{fig:sketch} visualizes the results of LAS-Diffusion and our methods. Although LAS-Diffusion generates plausible results which basically conform to the geometry of input sketch, it struggles to recover the fine details of 3D shapes such as the wheel hub as well as horizontal and vertical bars of chair. On the contrary, our method is capable of possessing better geometry quality and matching better with the input sketch.

\subsubsection{Texture Generation}
\begin{figure}[t]
  \includegraphics[width=0.9\columnwidth]{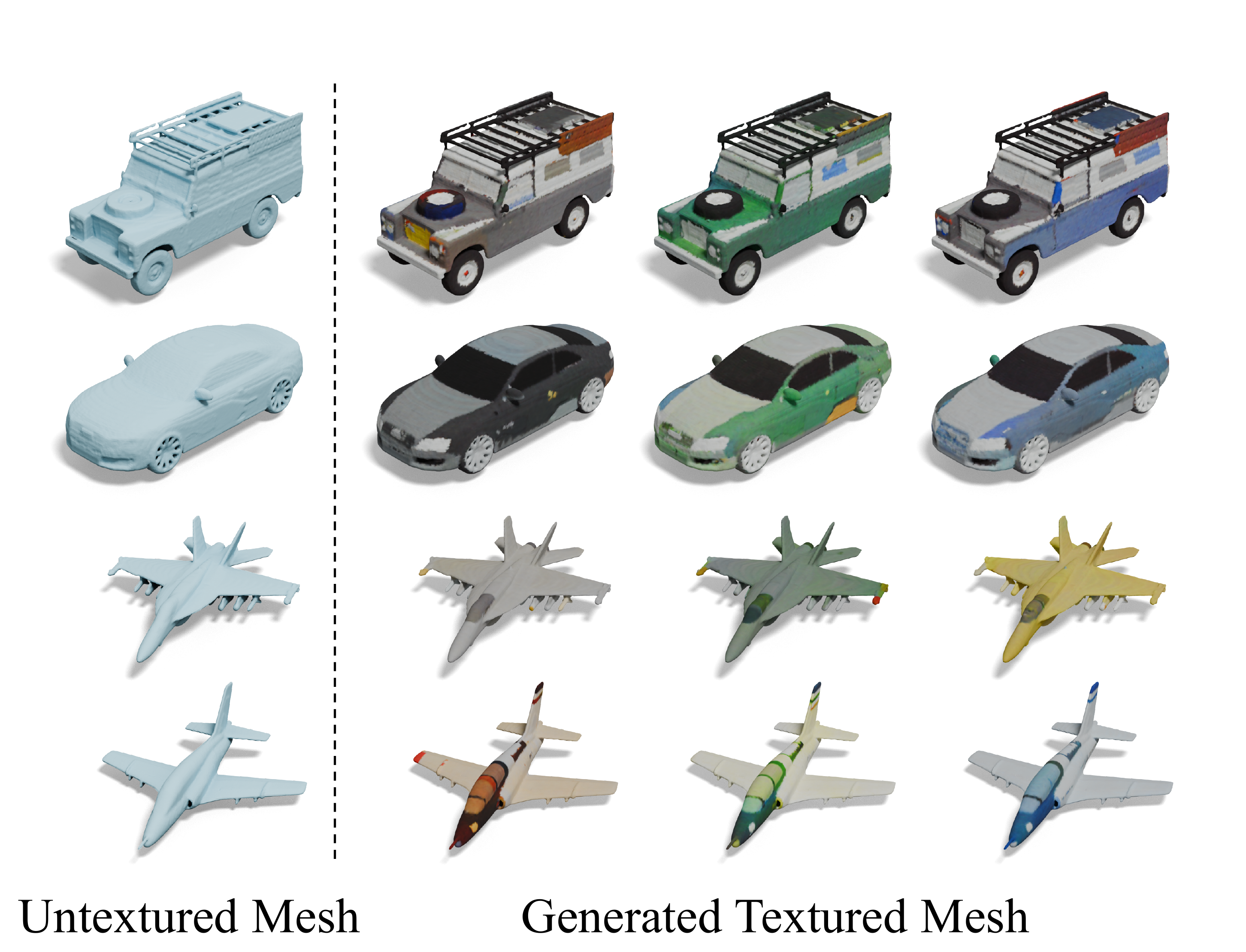}
  \caption{The performance of octree texture diffusion module which is capable of producing diverse texture map given a single input.}
  \label{fig:texture}
\end{figure}

Based on the trained OctFusion for textured shape generation, we can synthesize texture maps given a single input untextured 3D shape or the corresponding geometric latent features.
Given initial texture features randomly sampled from a Gaussian distribution, our OctFusion can progressively denoise the texture features to generate the resulting texture maps.
Fig.~\ref{fig:texture} shows the results of our method. It can be seen that our method is capable of generating high-quality and diverse texture maps that match well with the input 3D shapes.

\section{Conclusion} \label{sec:conclusion}

We propose OctFusion, a novel  diffusion model for 3D shape generation.
The key contributions of our OctFusion include an octree-based latent representation and a unified U-Net architecture for high-resolution 3D shape generation.
OctFusion generate high-quality 3D shapes with fine details and textures and can support various generative applications, such as text/sketch-conditioned generation.
OctFusion is currently lightweight and contains only $33.03M$ parameters.
We expect that our OctFusion can be easily scaled up and greatly improved if more data and computational resources are available.
And we will explore the possibility of training multiple stages of U-Net simultaneously in the future.


\section*{Acknowledgements}
This work was supported in part by National Key R\&D Program
of China 2022ZD0160801, Beijing Natural Science Foundation
No. 4244081, National Natural Science Foundation of China (Grant No.: 62372015), and Tencent AI Lab Rhino-Bird Focused Research Program.
We also thank the anonymous reviewers for their invaluable feedback.

\printbibliography    

\clearpage
\appendix
\section{Implementation Details}

\subsection{Textured Shape Generation}
This section introduces the implementation details for textured shape generation.
For data preparation, we use the same procedure in Section 4.1 to repair all the meshes in ShapeNet to obtain dense on-surface points and off-surface query points and their corresponding SDF values. 
Then, we reproject these points onto the surface of the original meshes and extract their corresponding RGB values to obtain colored input point clouds.
We build octrees with depth 8 (resolution $256^3$) with colored point clouds as the input of VAE.
The VAE has two separate decoders, and the latent feature dimension is set to 6 with 3 channels for geometry and another 3 channels for color.
Then, we train a two-stage OctFusion. Finally, we train an additional octree-based texture diffusion model to generate color latent code based on the geometry latent code to attain a textured 3D mesh followed by the decoder of extended VAE.

\subsection{Model Architecture}

\subsubsection{Octree-based VAE} 
\begin{figure}[t]
  \centering
  \includegraphics[width=\linewidth]{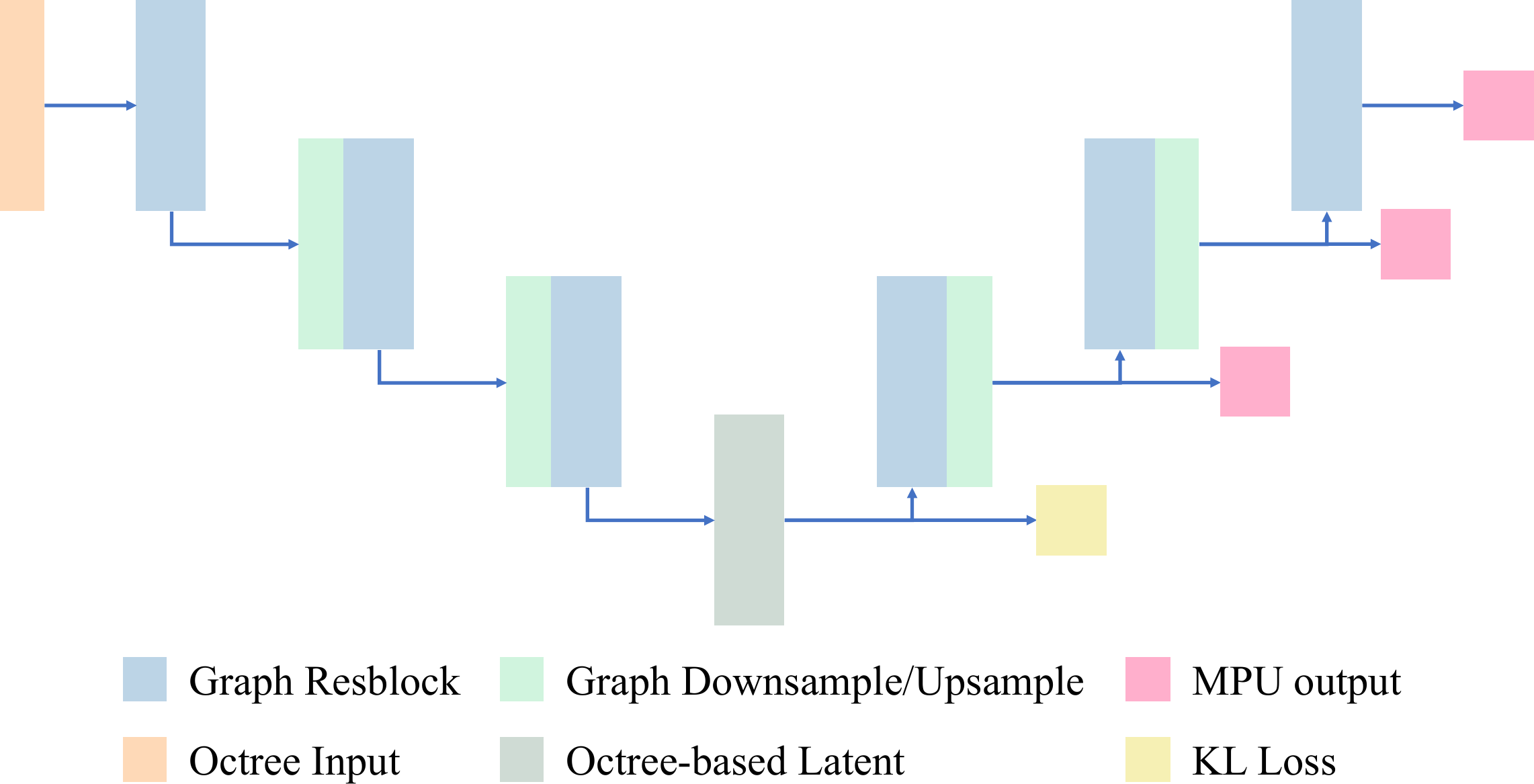}
  \caption{The network architecture of Octree-based VAE.}
  \label{fig:vae}
\end{figure}

The network architecture of the octree-based latent Variational Autoencoder (VAE), as depicted in Figure~\ref{fig:vae}, is constructed via dual-octree graph convolution network. For two-stage OctFusion model, the VAE has three hierarchical levels, corresponding to octree depths of 8, 7, and 6, with corresponding resolutions of  $256^3$, $128^3$, $64^3$. The feature dimensions are set to 24, 32 and 32 respectively.

\subsubsection{OctFusion} 
\begin{figure}[t]
  \centering
  \includegraphics[width=\linewidth]{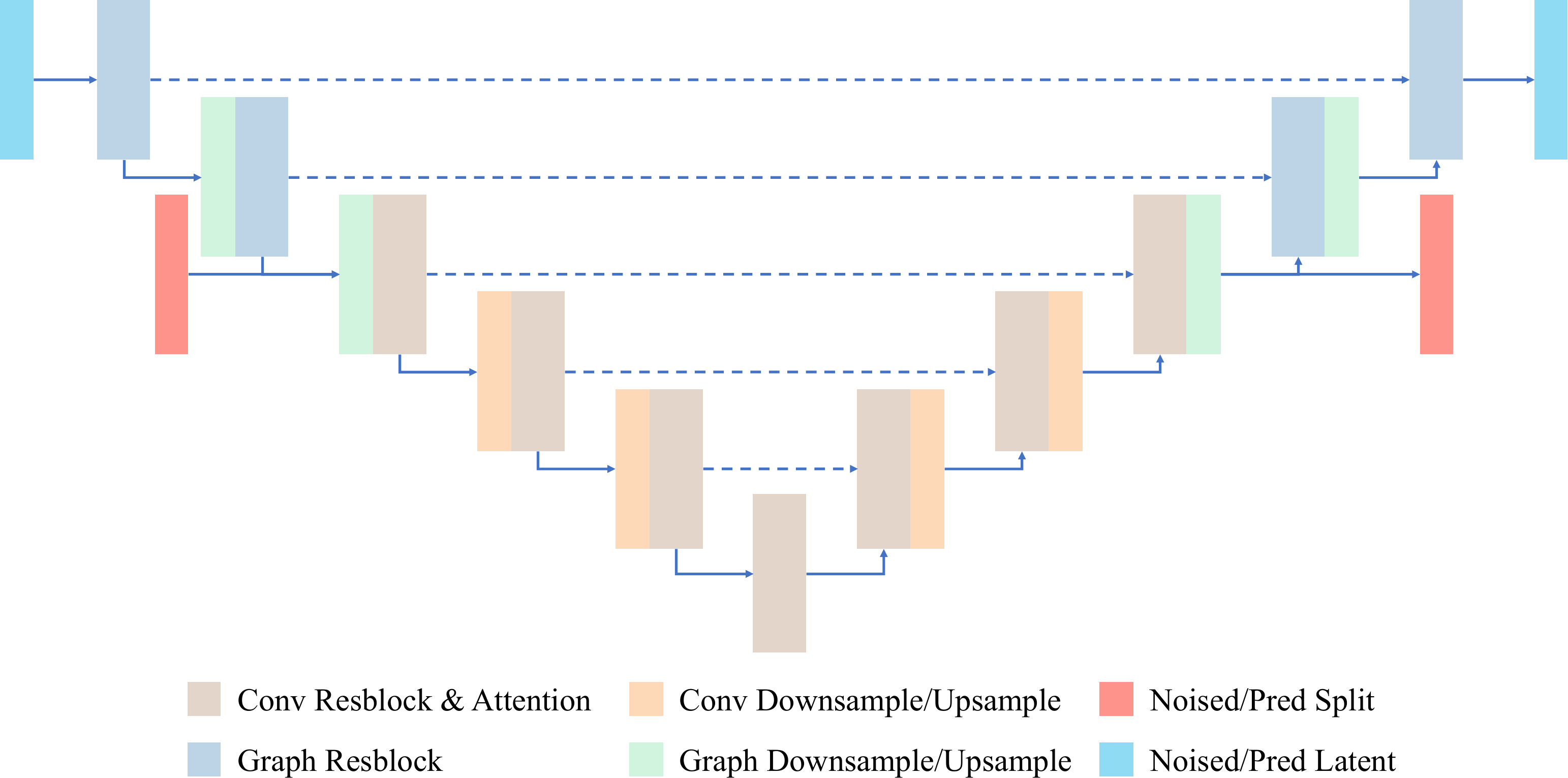}
  \caption{The network architecture of OctFusion U-Net.}
  \label{fig:unet_2t}
\end{figure}

We present the unified U-Net architecture of OctFusion in Fig.~\ref{fig:unet_2t}. The first stage, denoted as $\mathcal{F}_1$, is designed for generating the splitting signals, using convolutional neural network with residual connection and self-attention block. The U-Net in $\mathcal{F}_1$ is composed of three levels: $16^3$, $8^3$, $4^3$, each associated with model channels of 64, 128, and 256, respectively. 
For two-stage OctFusion model, the second stage $\mathcal{F}_2$ model is used for predicting clean latent features on each octree leaf node. The U-Net is constructed via dual octree graph convolution and has two levels $64^3$, $32^3$. The corresponding model channels are 128 and 256. At the bottom of $\mathcal{F}_2$ U-Net, the features are downsampled to $16^3$ and fed into $\mathcal{F}_1$.

\begin{figure}[t!]
  \centering
  \includegraphics[width=\linewidth]{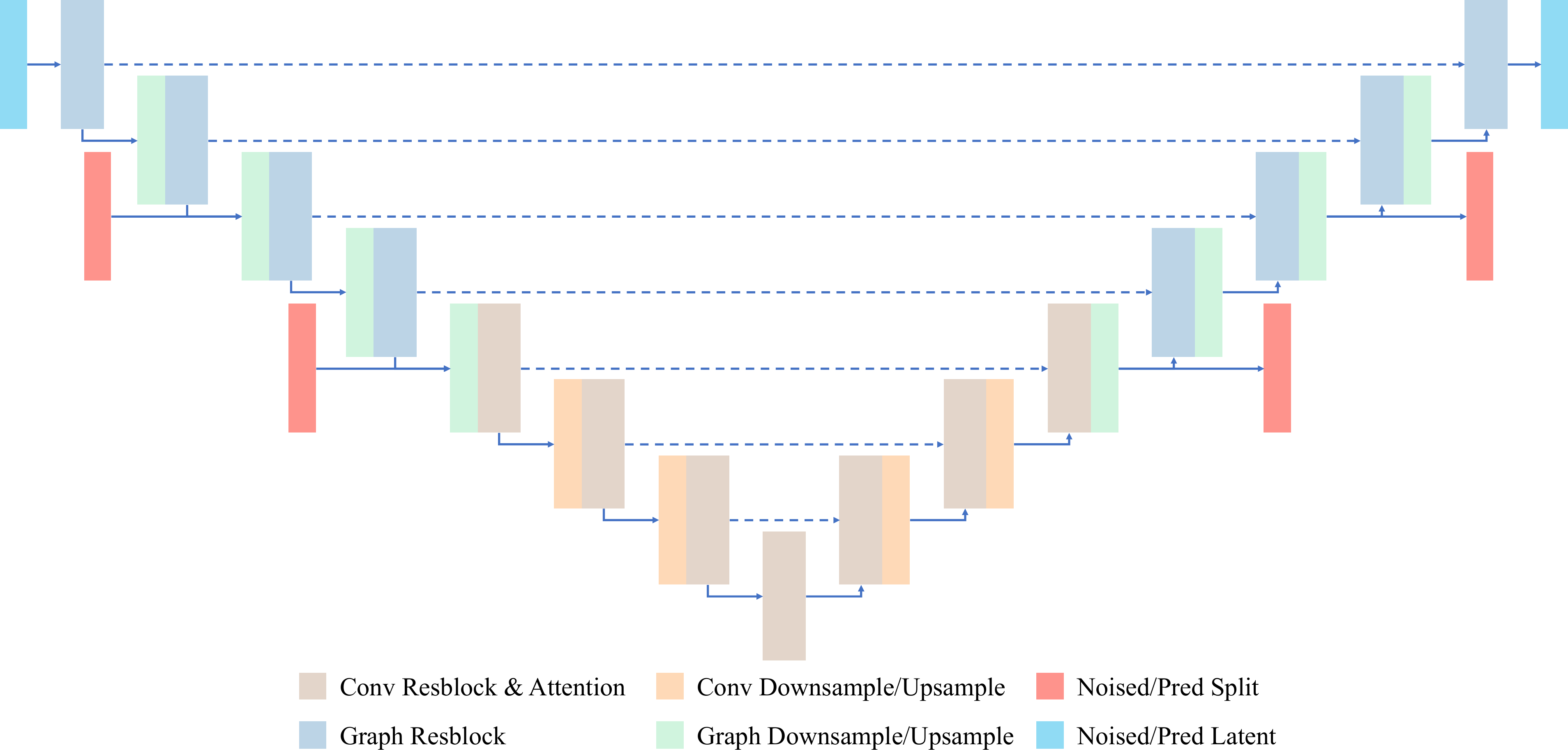}
  \caption{The network architecture of deeper OctFusion U-Net.}
  \label{fig:unet_3t}
\end{figure}

\begin{figure}[t!]
  \centering
  \includegraphics[width=\linewidth]{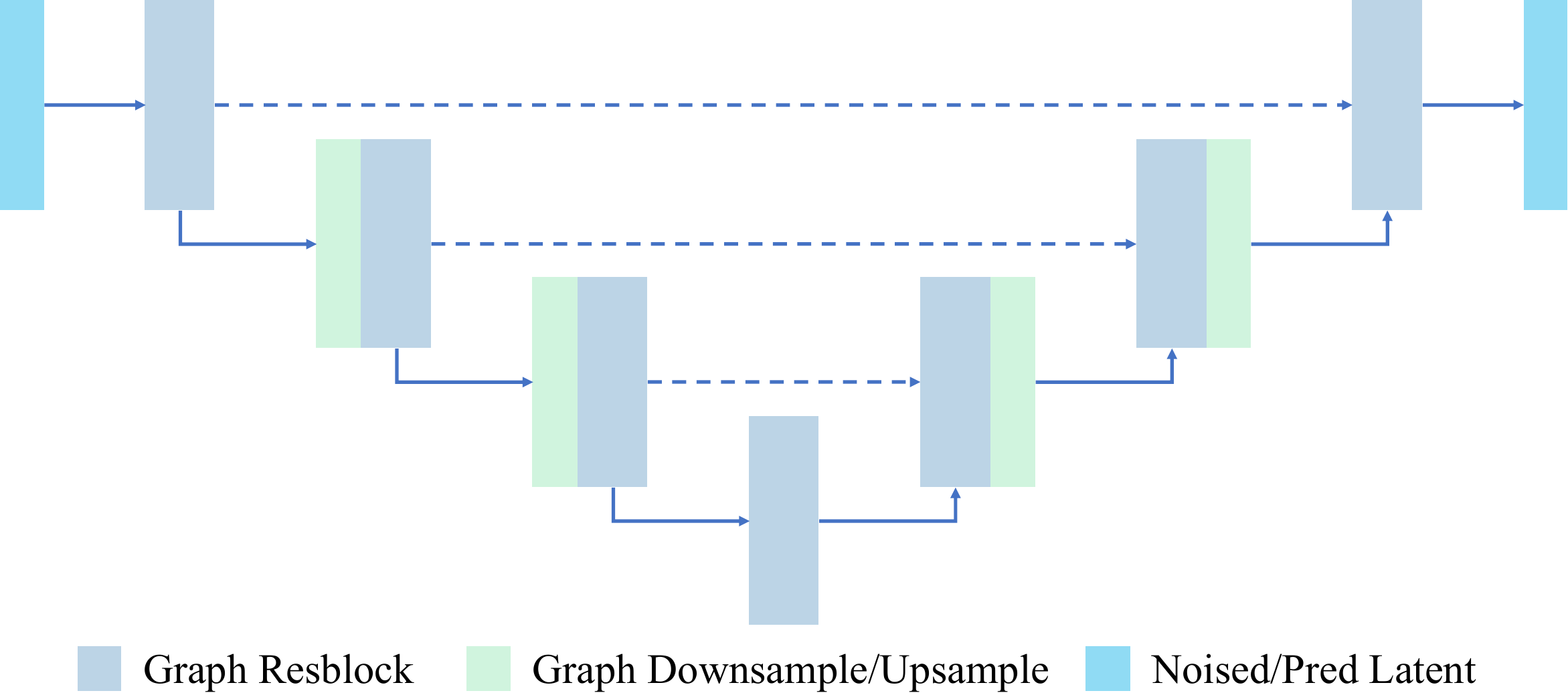}
  \caption{The network architecture of OctFusion U-Net for color generation.}
  \label{fig:unet_color}
\end{figure}

The deeper OctFusion (such as three-stage) is shown in Fig.~\ref{fig:unet_3t}. The $\mathcal{F}_1$ and $\mathcal{F}_2$ models is used for generating splitting signals and have the same network architecture mentioned above. The $\mathcal{F}_3$ predicts the clean latent code on the octree generated by $\mathcal{F}_1$ and $\mathcal{F}_2$. The U-Net of $\mathcal{F}_3$ has two levels $256^3$ and $128^3$, with associated model channels of 64 and 128. We also present the network architecture of octree-based texture diffusion model in Fig.~\ref{fig:unet_color}, which is used for generating color latent codes for existing untextured 3D shape. Our octree-based texture diffusion model has the same architecture as $\mathcal{F}_3$ but is not unified with lower stages.

\section{Metric Definition}

\subsubsection{Distance}
We begin by sampling points from the surfaces of both the generated mesh and the reference mesh in dataset, resulting in the point clouds denoted as $S_g$ and $S_r$, respectively. 
Distance between two point clouds can be evaluated by Chamfer Distance(CD) and Earth Mover's Distance (EMD). Chamfer Distance is a symmetric measure that calculates the average distance from points in $S_g$ to the nearest points in $S_r$, and vice versa. The formula for Chamfer Distance is given by:

\begin{equation}
    \text{CD}(S_g, S_r) = \sum_{x\in S_g}\min_{y\in S_r} ||x-y||^2_2 + \sum_{y\in S_r}\min_{x\in S_g}||x-y||^2_2.
    \label{CD}
\end{equation}

where $d(x, y)$ is the Euclidean distance between $X$ and $Y$.
Earth Mover's Distance can be treated of as the minimum transportation from one point cloud into another. The formula of Earth Mover's Distance (EMD) is defined as:
\begin{equation}
    \text{EMD}(S_g, S_r) = \min_{\phi: S_g \rightarrow S_r} \sum_{X \in S_g} \| X - \phi(X) \|_2
    \label{EMD}
\end{equation}
where $\phi$ is a bijection.

\subsubsection{Coverage (COV)}
Coverage is calculated as the fraction of point clouds in the reference set that are matched to at least one point cloud in the generated set. For each point cloud in the generated set, its near neighbor in the reference set is marked as a match:
\begin{equation}
    COV(S_g, S_r) = \frac{| \{ \text{argmin}_{Y \in S_r} D(X, Y) | X \in S_g \} |}{| S_r |}
    \label{COV}
\end{equation}
where $D(\cdot,\cdot)$ can be either CD or EMD. A high coverage score indicated that most of reference set is roughly represented within generated set.

\subsubsection{Minimum Matching Distance(MMD)} 
Minimum Matching Distance is proposed to
complement coverage as a metric that measures quality.
For each point cloud in the reference set, the distance to its nearest neighbor in the generated set is computed and averaged:
\begin{equation}
    \text{MMD}(S_g, S_r) = \frac{1}{|S_g|} \sum_{Y \in S_r} \min_{X \in S_g} D(X, Y)
    \label{MMD}
\end{equation}
where $D(\cdot, \cdot)$ can be either CD or EMD.
Since MMD relies directly on the distances of the matching, it correlates well with how faithful (with respect
to the reference set) the elements of generated set are.

\subsubsection{1-NNA} The 1-Nearest Neighbor Assignment (1-NNA) metric evaluates the classification accuracy when employing the nearest neighbor criterion under distance measure $D$ to indicate whether a point cloud is synthetic or not. Ideally, if the generated point cloud closely mirrors the distribution of the reference set, the classification accuracy should be around 50\%. The formula for 1-NNA is defined as follows.
\begin{equation}
    \text{1-NNA}(S_g, S_r) = \frac{\sum_{X \in S_g} \mathbb{1}[N_x \in S_g] + \sum_{Y \in S_r} \mathbb{1}[N_y \in S_r]}{|S_g| + |S_r|}
    \label{1-NNA}
\end{equation}
where $N_X$ is the closest point cloud to $X$ under distance metric $D(\cdot, \cdot)$, and $\mathbb{1}[\cdot]$ is the indicator function.

\subsubsection{shading-image-based FID} \textit{shading-image-based FID} is a more robust measure for evaluating both the quality and diversity of generated shapes. To compute the FID metric, each generated shape is rendered from 20 uniformly distributed viewpoints around the shape. These rendered shading images are then used to calculate the FID scores on the rendered image set of the original training dataset. The formula for FID is defined as follows.
\begin{equation}
    \text{FID} = \frac{1}{20} \sum_{i=1}^{20} \|\mu_g^i - \mu_r^i\|^2 + \text{Tr}(\Sigma_g^i + \Sigma_r^i - 2(\Sigma_g^i \Sigma_r^i)^{1/2})
\end{equation}
where $\mu^i$ and $\Sigma^i$ denote the mean and covariance of the $i$-th view's shading images.

\section{More Results on ShapeNet}
We present more unconditional generation results on ShapeNet category \texttt{chair}, \texttt{table}, \texttt{airplane}, \texttt{car}, and \texttt{rifle} in the following pages. These results demonstrate the quality and diversity of our proposed  OctFusion.

\begin{figure*}[t!]
  \centering
  \includegraphics[width=\textwidth]{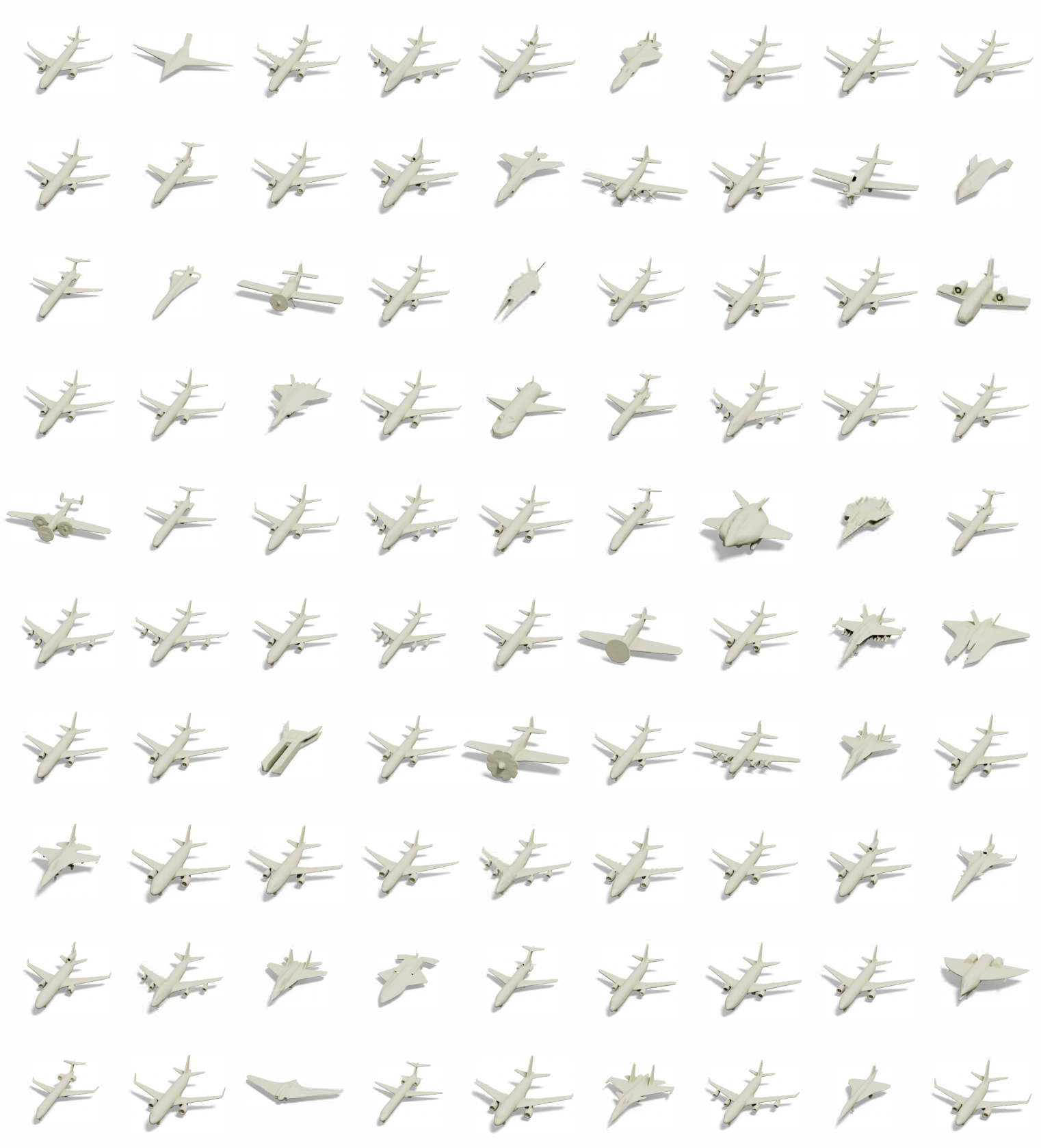}
  \caption{More generative results on airplane}
  \label{fig:multiple-airplane1}
\end{figure*}

\begin{figure*}[t!]
  \centering
  \includegraphics[width=\textwidth]{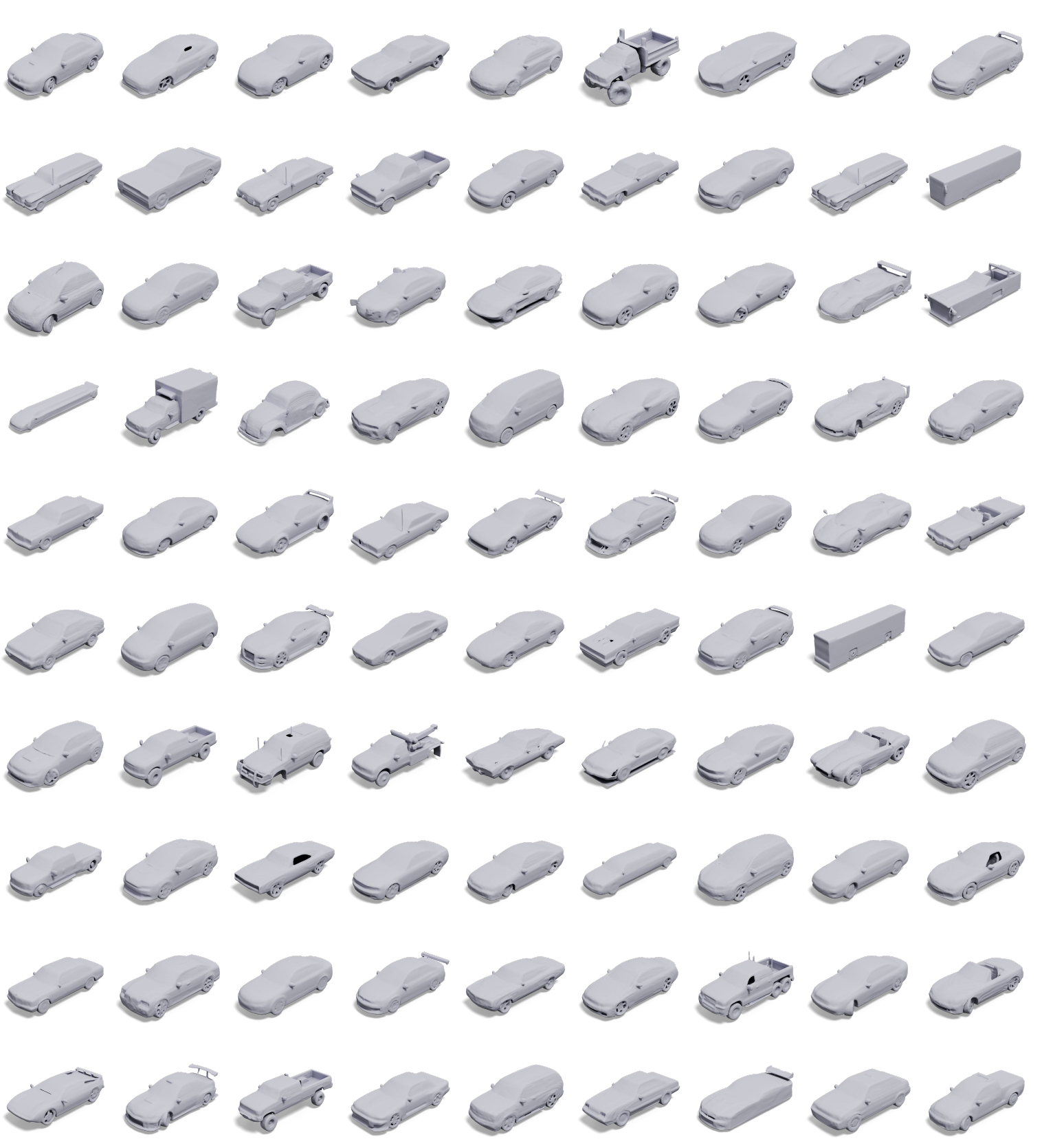}
  \caption{More generative results on car}
  \label{fig:multiple-car1}
\end{figure*}

\begin{figure*}[t!]
  \centering
  \includegraphics[width=\textwidth]{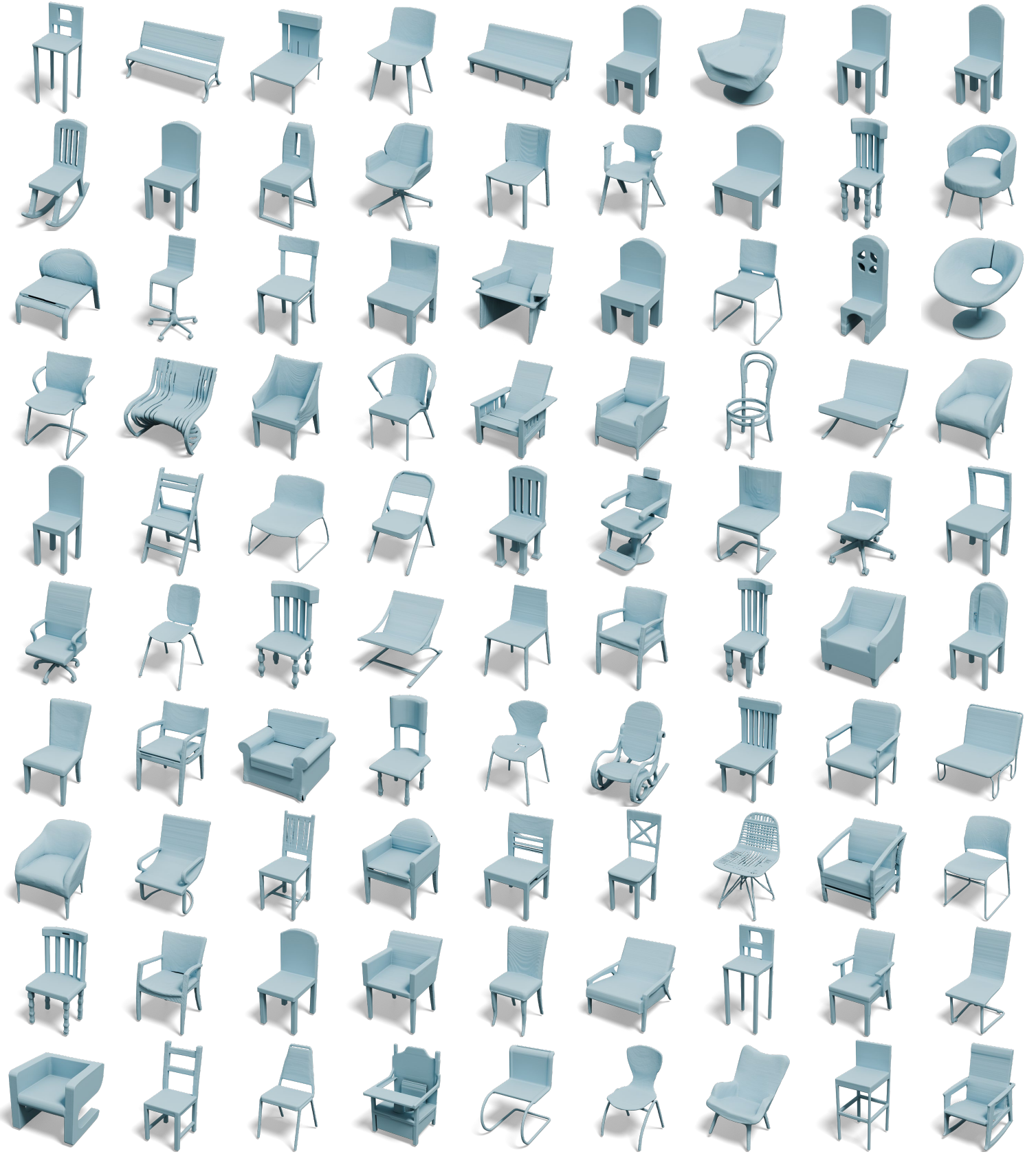}
  \caption{More generative results on chair}
  \label{fig:multiple-chair1}
\end{figure*}

\begin{figure*}[t!]
  \centering
  \includegraphics[width=\textwidth]{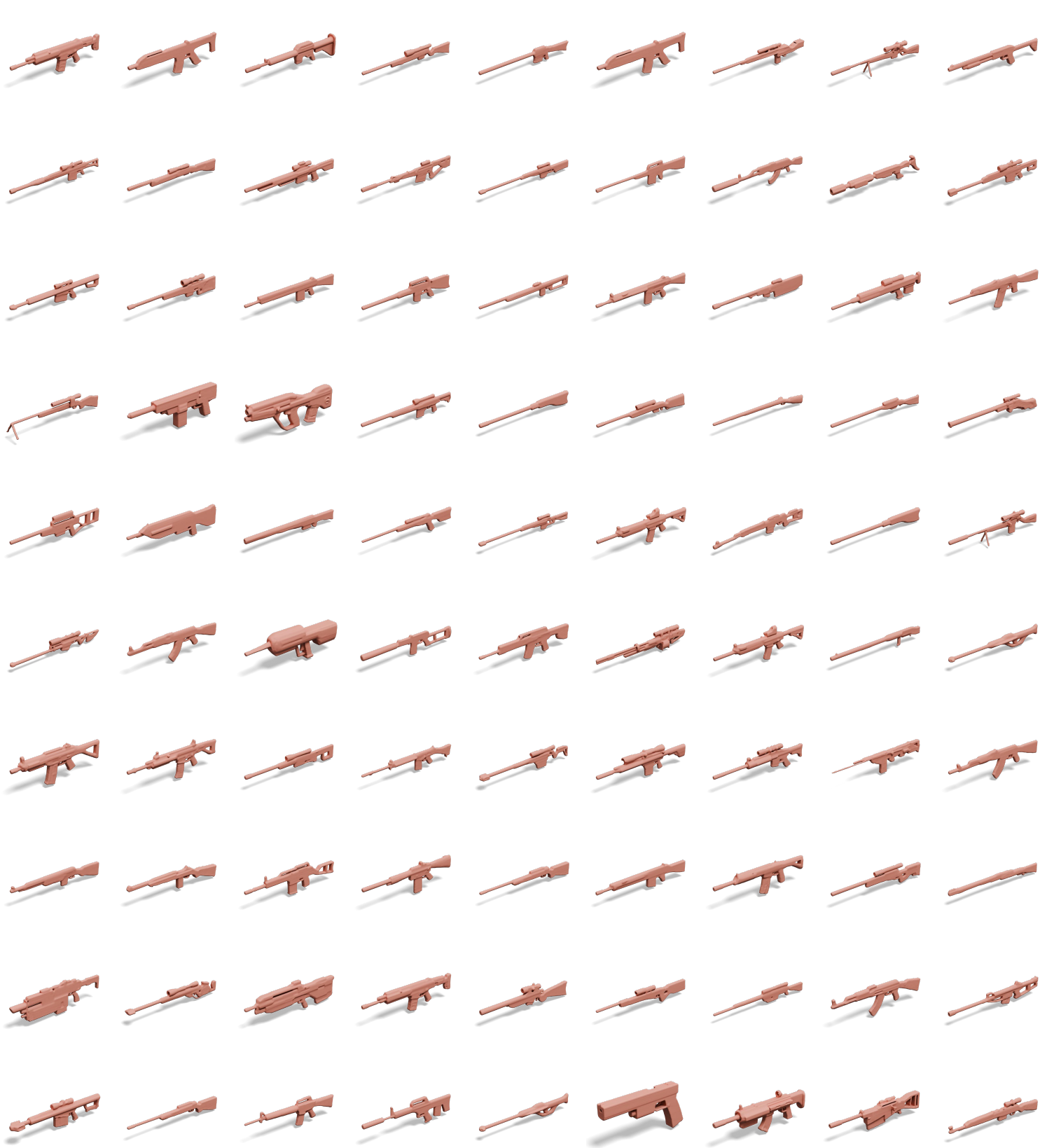}
  \caption{More generative results on rifle}
  \label{fig:multiple-rifle1}
\end{figure*}

\begin{figure*}[t!]
  \centering
  \includegraphics[width=\textwidth]{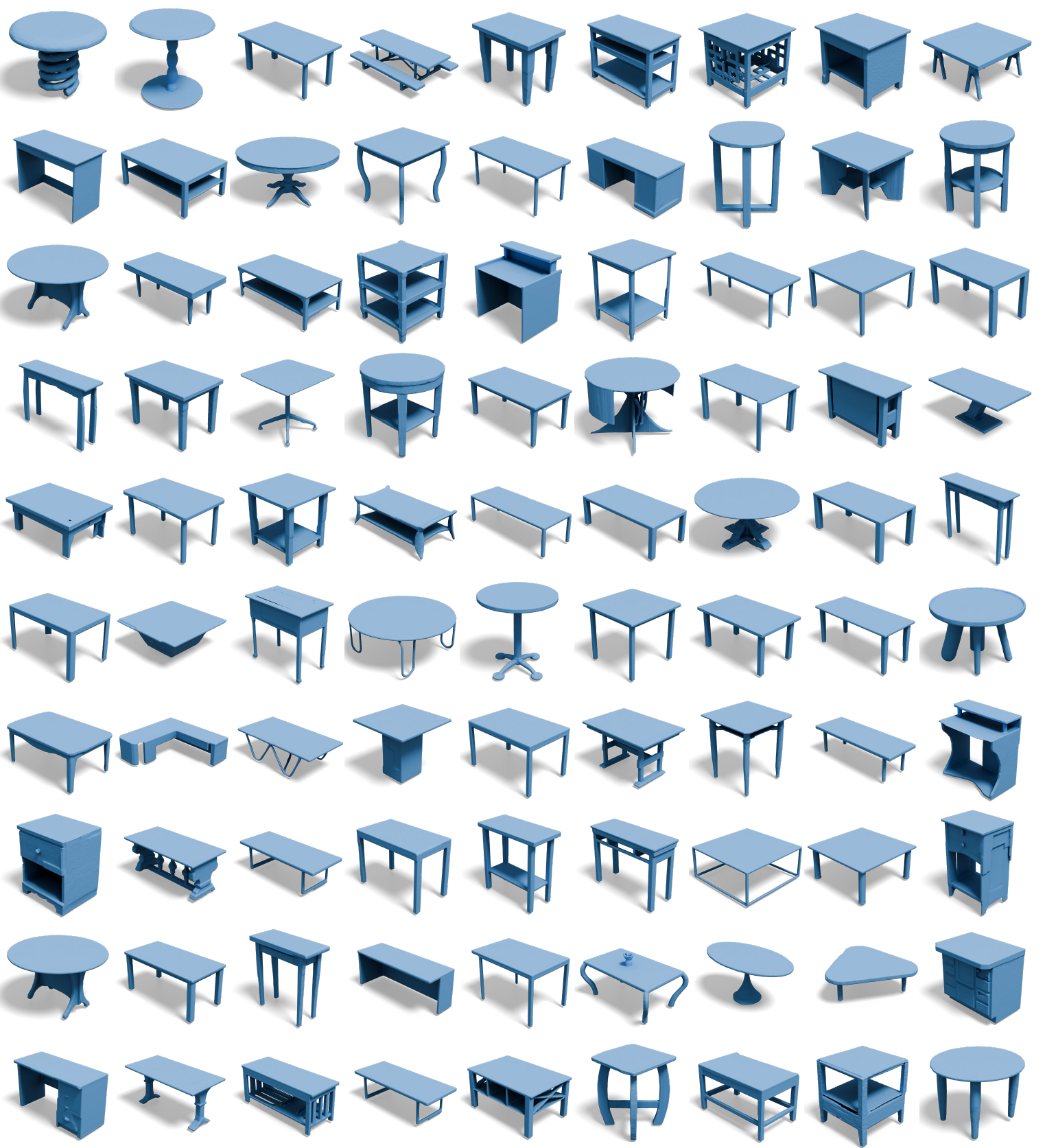}
  \caption{More generative results on table}
  \label{fig:multiple-table1}
\end{figure*}

\end{document}